\documentclass[12pt]{article}

\makeatletter
\def\newleaf{\newpage
\newcount\tmp
\tmp=\c@page
\divide\tmp by 2
\multiply\tmp by 2
\ifnum\c@page=\tmp
~\newpage
\fi
}
\makeatother

\expandafter\ifx\csname proofnewpage\endcsname\relax

\fi

\def\color[#1]#2{}

\long\def\nop#1{}

\def\comment{\edef\cps{\the\parskip} \parskip=0.5cm \begingroup \tt}

\expandafter\ifx\csname shortcite\endcsname\relax
\let\shortcite=\cite
\fi

\newbox\current

\long\def\plframebox#1{
\setbox\current\vbox{#1}		

\vbox to \ht\current {\hrule\vss
\hbox to \wd\current {%
\vrule \hss\box\current\hss \vrule}
\vss\hrule }
}

\long\def\eatpar#1{%
\ifx#1\par                      
\let\nextmove=\eatpar           
\else
\let\nextmove=#1
\fi
\noexpand\nextmove
}

\def\modifymargins#1#2{
\newdimen\addtoh
\newdimen\addtow
\addtoh=#1
\addtow=#2

\advance\topmargin by -\addtoh
\multiply\addtoh by 2
\advance\textheight by \addtoh

\advance\oddsidemargin by -\addtow
\advance\evensidemargin by -\addtow
\multiply\addtow by 2
\advance\textwidth by \addtow
}

\begingroup
\catcode`\~=11
\gdef\centertilde#1{\lower #1pt\hbox{~}}
\endgroup

\newcount\currenttime
\newcount\hour
\newcount\minute

\def\printtime{%
\currenttime=\time
\hour=\currenttime
\divide\hour by 60
\minute=-\hour
\multiply\minute by 60
\advance\minute by \currenttime
\the\hour:\ifnum\minute<10 0\fi\the\minute
}

\begingroup
\makeatletter
\global\let\@@date=\@date
\gdef\@date{\@@date\ --- \printtime}
\endgroup

\def\oggi{\number\day\space 
\ifcase\month\or
Gennaio\or Febbraio\or Marzo\or Aprile\or Maggio\or Giugno\or
Luglio\or Agosto\or Settembre\or Ottobre\or Novembre\or Dicembre\fi
\space \number\year}

\newcounter{rmexample}

\def\proof{\noindent {\sl Proof.\ \ }}

\def\qed{\hfill{\boxit{}}
  \ifdim\lastskip<\medskipamount \removelastskip\penalty55\medskip\fi}
\def\qedn#1{\hfill{\boxit{}$_#1$}
  \ifdim\lastskip<\medskipamount \removelastskip\penalty55\medskip\fi}
\long\def\boxit#1{\vbox{\hrule\hbox{\vrule\kern3pt
                  \vbox{\kern3pt#1\kern3pt}\kern3pt\vrule}\hrule}}

\def\bh#1{\if#1{}{\rm BH}\else\mbox{BH$_{#1}$}\fi}

\def\profont{\sf}

\def\x3c{{\profont x3c}}

\def\possnewtheorem#1#2{
\expandafter\ifx\csname #1\endcsname\relax
\newtheorem{#1}{#2}
\fi
}

\def\possnewtheoremthree#1[#2]#3{
\expandafter\ifx\csname #1\endcsname\relax
\newtheorem{#1}[#2]{#3}
\fi
}

\possnewtheorem{theorem}{Theorem}
\possnewtheorem{corollary}{Corollary}
\possnewtheorem{lemma}{Lemma}
\possnewtheoremthree{proposition}[theorem]{Proposition}
\possnewtheorem{definition}{Definition}
\possnewtheorem{question}{Question}
\possnewtheorem{example}{Example}
\possnewtheorem{nontheorem}{Counterexample}
\possnewtheorem{property}{Property}
\possnewtheorem{assumption}{Assumption}
\possnewtheorem{conjecture}{Conjecture}
\possnewtheorem{notation}{Notation}
\newenvironment{theorem*}[1]{{\noindent \bf Theorem~#1}\begin{it}}{\end{it}\

}

\modifymargins{80pt}{40pt}

\possnewtheorem{algorithm}{Algorithm}

\iffalse
\def\frac#1#2{#1/#2}
\makeatletter
\def\@ttyfig#1#2{\def\specialtext{\special{txt:#2}}\specialtext\egroup}
\nop{
\let\ttytex\ttyfig
\makeatother
\else
\def\ttytex#1{#1\nop}			
\fi

\title{Belief Merging by Source Reliability Assessment}
\author{%
Paolo Liberatore%
\thanks{Dipartimento di Ingegneria Informatica, Automatica e Gestionale,
Sapienza University of Rome,
Via Ariosto 25, 00185, Rome, Italy.
Email: {\tt paolo@liberatore.org}}
}
\date{}

\begin{document}

\maketitle

\begin{abstract}

Merging beliefs requires the plausibility of the sources of the information to
be merged. They are typically assumed equally reliable in lack of hints
indicating otherwise~\cite{koni-pere-11}; yet, a recent line of research spun
from the idea of deriving this information from the revision process itself. In
particular, the history of previous revisions~\cite{boot-nitt-08,libe-15} and
previous merging examples~\cite{libe-16} provide information for performing
subsequent mergings.

Yet, no examples or previous revisions may be available. In spite of the
apparent lack of information, something can still be inferred by a
try-and-check approach: a relative reliability ordering is assumed, the merging
process is performed based on it, and the result is compared with the original
information. The outcome of this check may be incoherent with the initial
assumption, like when a completely reliable source is rejected some of the
information it provided. In such cases, the reliability ordering assumed in the
first place can be excluded from consideration. The first theorem of this
article proves that such a scenario is indeed possible. Other results are
obtained under various definition of reliability and merging.

\end{abstract}

\section{Introduction}
\label{introduction}

Between November 2005 and September 2006 Wikipedia had an article about
Porchesia, a 300,000-inhabitant island in the Mediterranean sea~\cite{porc-06}.
No such place was ever mentioned in the article on the Mediterranean sea or
those of Europe, Asia or Africa. It took ten months for the article to be find
out an hoax and removed.

Wikipedia accepts information from any user who want to provide some, even
anonymously; no prior barrier exists, not even for article creation. This was
the reason of its success: a previous project of creating an on-line
encyclopedia from reputed experts failed~\cite{wile-gurr-09}. At the same time,
it opens the door to wrong and malicious information. This is not limited to
Wikipedia. A number of other successful sites work in a similar way: computer
programmers rely on Stack Overflow~\cite{baru-thom-hass-12}, information on
many topics can be found on Stack Exchange~\cite{posn-etal-12} and Yahoo
Answers~\cite{adam-etal-08}. Everyone can provide information; checking is done
later: with article removal or correction on Wikipedia, with a measure of trust
on other Internet fora.

The Porchesia article is an extreme example, but shows the difficulty of
manually checking a large amount of information coming from unknown sources,
even by a large and dedicated community~\cite{vieg-etal-07}. A large island
should have been at least mentioned in related articles, such as that about the
Mediterranean sea, but it was not. But these articles also come from users of
unknown reliability. The conflict is between sources of unknown reliability.

Unknown does not mean the same for all, as commonly assumed in the theory of
belief merging~%
\cite{koni-pere-11,libe-scha-98-b,ever-koni-marq-15,%
cevo-14,herz-pozo-schw-14,dalf-15}.
The user who created the hoax article is not to be trusted much on other
additions.

Assuming a prior assessment of the sources would be unrealistic. Not only web
sites such as Wikipedia and Stack Overflow rely on new users providing useful
information. Even in more controlled scenarios such as database integration
assigning priority to sources is only one of several value conflict
strategies~\cite{naum-etal-06}, the others including ``ask the user'', ``take
the most used value'', ``if the value is numeric, use the average''. Some works
in belief revision try to derive such a preference from the previous history of
revisions~\cite{boot-nitt-08,libe-15} or from examples~\cite{libe-16}. Neither
is assumed known in this article.

The scenario considered in this article comprises some sources of information,
each providing a number of propositional formulae. In the simplest case of the
usual definition of merging, each source gives exactly one formula. More
generally, several pieces of information may come from the same source: a
sensor produces several readings, a database can be queried several times, etc.

\begingroup\makeatletter\ifx\SetFigFont\undefined%
\gdef\SetFigFont#1#2#3#4#5{%
  \reset@font\fontsize{#1}{#2pt}%
  \fontfamily{#3}\fontseries{#4}\fontshape{#5}%
  \selectfont}%
\fi\endgroup%

\begin{figure}[ht]
\ttytex{
\begin{center}
\hfill\break
\setlength{\unitlength}{5000sp}%
\begingroup\makeatletter\ifx\SetFigFont\undefined%
\gdef\SetFigFont#1#2#3#4#5{%
  \reset@font\fontsize{#1}{#2pt}%
  \fontfamily{#3}\fontseries{#4}\fontshape{#5}%
  \selectfont}%
\fi\endgroup%
\begin{picture}(3264,1470)(3319,-4306)
\thinlines
{\color[rgb]{0,0,0}\put(6031,-3211){\line(-2, 1){720}}
\put(5311,-2851){\line( 0,-1){1440}}
\put(5311,-4291){\line( 2, 1){720}}
\put(6031,-3931){\line( 0, 1){720}}
}%
{\color[rgb]{0,0,0}\put(4141,-3031){\vector( 1, 0){1170}}
}%
{\color[rgb]{0,0,0}\put(4141,-4111){\vector( 1, 0){1170}}
}%
{\color[rgb]{0,0,0}\put(4141,-3571){\vector( 1, 0){1170}}
}%
{\color[rgb]{0,0,0}\put(6031,-3571){\vector( 1, 0){540}}
}%
{\color[rgb]{0,0,0}\put(3331,-3211){\framebox(810,360){}}
}%
{\color[rgb]{0,0,0}\put(3331,-3751){\framebox(810,360){}}
}%
{\color[rgb]{0,0,0}\put(3331,-4291){\framebox(810,360){}}
}%
\put(5671,-3616){\makebox(0,0)[b]{\smash{{\SetFigFont{12}{24.0}{\rmdefault}{\mddefault}{\updefault}{\color[rgb]{0,0,0}$merger$}%
}}}}
\put(3736,-4156){\makebox(0,0)[b]{\smash{{\SetFigFont{12}{24.0}{\rmdefault}{\mddefault}{\updefault}{\color[rgb]{0,0,0}$source 3$}%
}}}}
\put(3736,-3616){\makebox(0,0)[b]{\smash{{\SetFigFont{12}{24.0}{\rmdefault}{\mddefault}{\updefault}{\color[rgb]{0,0,0}$source 2$}%
}}}}
\put(3781,-3076){\makebox(0,0)[b]{\smash{{\SetFigFont{12}{24.0}{\rmdefault}{\mddefault}{\updefault}{\color[rgb]{0,0,0}$source 1$}%
}}}}
\put(4996,-4291){\makebox(0,0)[b]{\smash{{\SetFigFont{12}{24.0}{\rmdefault}{\mddefault}{\updefault}{\color[rgb]{0,0,0}$kb2$}%
}}}}
\put(4681,-4066){\makebox(0,0)[b]{\smash{{\SetFigFont{12}{24.0}{\rmdefault}{\mddefault}{\updefault}{\color[rgb]{0,0,0}$kb6$}%
}}}}
\put(4456,-3526){\makebox(0,0)[b]{\smash{{\SetFigFont{12}{24.0}{\rmdefault}{\mddefault}{\updefault}{\color[rgb]{0,0,0}$kb7$}%
}}}}
\put(4906,-3526){\makebox(0,0)[b]{\smash{{\SetFigFont{12}{24.0}{\rmdefault}{\mddefault}{\updefault}{\color[rgb]{0,0,0}$kb4$}%
}}}}
\put(4726,-3211){\makebox(0,0)[b]{\smash{{\SetFigFont{12}{24.0}{\rmdefault}{\mddefault}{\updefault}{\color[rgb]{0,0,0}$kb3$}%
}}}}
\put(4366,-2986){\makebox(0,0)[b]{\smash{{\SetFigFont{12}{24.0}{\rmdefault}{\mddefault}{\updefault}{\color[rgb]{0,0,0}$kb5$}%
}}}}
\put(5086,-2986){\makebox(0,0)[b]{\smash{{\SetFigFont{12}{24.0}{\rmdefault}{\mddefault}{\updefault}{\color[rgb]{0,0,0}$kb1$}%
}}}}
\put(4321,-4291){\makebox(0,0)[b]{\smash{{\SetFigFont{12}{24.0}{\rmdefault}{\mddefault}{\updefault}{\color[rgb]{0,0,0}$kb8$}%
}}}}
\put(6526,-3481){\makebox(0,0)[b]{\smash{{\SetFigFont{12}{24.0}{\rmdefault}{\mddefault}{\updefault}{\color[rgb]{0,0,0}$kb$}%
}}}}
\end{picture}%
\end{center}
}{
                       +---
         kb5 kb3 kb1   |   ---
source1 -------------> |      ---+
                       |         |
           kb7  kb4    |         |
source2 -------------> |  merger | -----> kb
                       |         |
         kb8 kb6 kb2   |         |
source3 -------------> |      ---+
                       |   ---
                       +---
}
\label{multiple-sources}
\caption{Multiple sources providing formulae to be merged}
\end{figure}
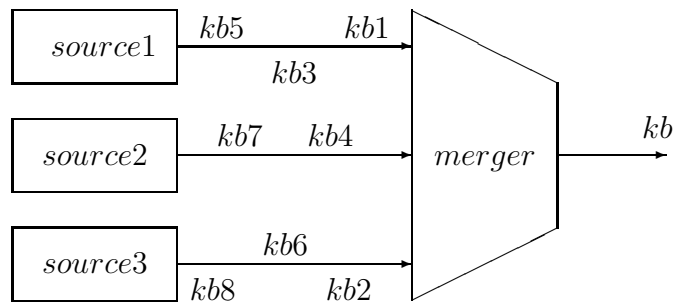

An example of such settings is in the Figure~\ref{multiple-sources}: $kb_1$,
$kb_3$ and $kb_5$ come from the first source, $kb_4$ and $kb_7$ from the
second, $kb_2$, $kb_6$ and $kb_8$ from the third. The sources may be
differently reliable. This affects the result of merging: if the first source
is more reliable than the third and $kb_2$ is the exact opposite of $kb_1$,
then $kb_1$ should be accepted and $kb_2$ rejected. In general, the merger
gives preference to formulae coming from reliable sources. Reliability is
attached to the sources, and is therefore the same for all formulae provided by
the same source. The case of different reliability in the same sources is
discussed separately.

The relative reliability of the sources may not be easy to obtain, as witnessed
by the large amount of work done assuming equal reliability~%
\cite{koni-pere-11,libe-scha-98-b,ever-koni-marq-15,%
cevo-14,herz-pozo-schw-14,dalf-15}. A recent line of research has tried to
address this problem by deriving this kind of metainformation from the merging
process itself: Booth and Meyer and Liberatore~\cite{boot-nitt-08,libe-15} used
the history of previous revisions to perform the following;
Liberatore~\cite{libe-16} exploited merging examples.

In lack of an explicit measure of source reliability, a previous history or a
set of merging example, the only sensible choice may seem to assume that all
formulae are equally reliable. Yet, something more can be said even in this
case from the very definition of reliability.

Reliability tells how much each formula affects the result of merging. If the
most reliable source says $x$ and the least says $\neg x$, the result should
imply $x$ or at least be consistent with it. Otherwise, merging did not really
reflect the considered reliability. In the other way around, the given
reliability was not coherent with merging.

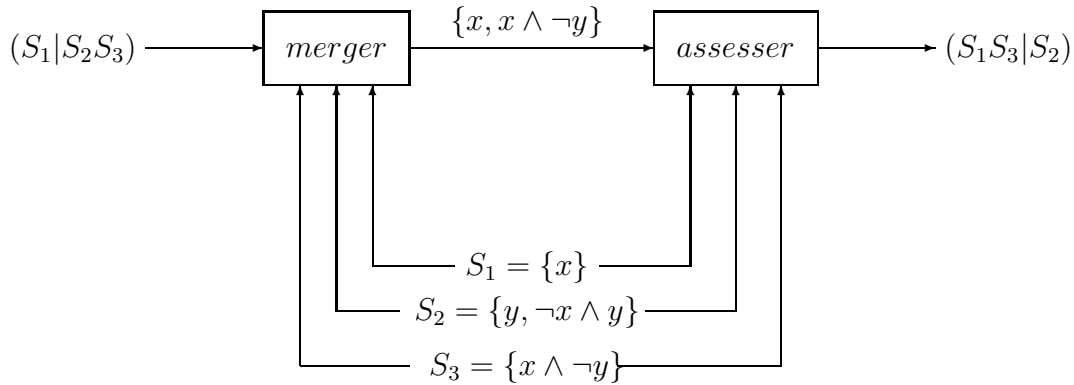
\begin{figure}[ht]
\ttytex{
\begin{center}
\hfill\break
\setlength{\unitlength}{5000sp}%
\begingroup\makeatletter\ifx\SetFigFont\undefined%
\gdef\SetFigFont#1#2#3#4#5{%
  \reset@font\fontsize{#1}{#2pt}%
  \fontfamily{#3}\fontseries{#4}\fontshape{#5}%
  \selectfont}%
\fi\endgroup%
\begin{picture}(4035,1938)(3361,-5080)
\thinlines
{\color[rgb]{0,0,0}\put(4006,-3571){\framebox(720,360){}}
}%
{\color[rgb]{0,0,0}\put(5941,-3571){\framebox(810,360){}}
}%
{\color[rgb]{0,0,0}\put(4726,-3391){\vector( 1, 0){1215}}
}%
{\color[rgb]{0,0,0}\put(3421,-3391){\vector( 1, 0){585}}
}%
{\color[rgb]{0,0,0}\put(6751,-3391){\vector( 1, 0){585}}
}%
{\color[rgb]{0,0,0}\put(5671,-4471){\line( 1, 0){450}}
\put(6121,-4471){\vector( 0, 1){900}}
}%
{\color[rgb]{0,0,0}\put(5761,-4966){\line( 1, 0){810}}
\put(6571,-4966){\vector( 0, 1){1395}}
}%
{\color[rgb]{0,0,0}\put(5896,-4696){\line( 1, 0){450}}
\put(6346,-4696){\vector( 0, 1){1125}}
}%
{\color[rgb]{0,0,0}\put(4951,-4471){\line(-1, 0){405}}
\put(4546,-4471){\vector( 0, 1){900}}
}%
{\color[rgb]{0,0,0}\put(4681,-4696){\line(-1, 0){315}}
\put(4366,-4696){\vector( 0, 1){1125}}
}%
{\color[rgb]{0,0,0}\put(4726,-4966){\line(-1, 0){540}}
\put(4186,-4966){\vector( 0, 1){1395}}
}%
\put(4366,-3436){\makebox(0,0)[b]{\smash{{\SetFigFont{12}{24.0}{\rmdefault}{\mddefault}{\updefault}{\color[rgb]{0,0,0}$merger$}%
}}}}
\put(6346,-3436){\makebox(0,0)[b]{\smash{{\SetFigFont{12}{24.0}{\rmdefault}{\mddefault}{\updefault}{\color[rgb]{0,0,0}$assesser$}%
}}}}
\put(3376,-3436){\makebox(0,0)[rb]{\smash{{\SetFigFont{12}{24.0}{\rmdefault}{\mddefault}{\updefault}{\color[rgb]{0,0,0}$(S_1|S_2S_3)$}%
}}}}
\put(5311,-3301){\makebox(0,0)[b]{\smash{{\SetFigFont{12}{24.0}{\rmdefault}{\mddefault}{\updefault}{\color[rgb]{0,0,0}$\{x, x \wedge \neg y\}$}%
}}}}
\put(5311,-4741){\makebox(0,0)[b]{\smash{{\SetFigFont{12}{24.0}{\rmdefault}{\mddefault}{\updefault}{\color[rgb]{0,0,0}$S_2=\{y, \neg x \wedge y\}$}%
}}}}
\put(5311,-4516){\makebox(0,0)[b]{\smash{{\SetFigFont{12}{24.0}{\rmdefault}{\mddefault}{\updefault}{\color[rgb]{0,0,0}$S_1=\{x\}$}%
}}}}
\put(5311,-5011){\makebox(0,0)[b]{\smash{{\SetFigFont{12}{24.0}{\rmdefault}{\mddefault}{\updefault}{\color[rgb]{0,0,0}$S_3=\{x \wedge \neg y\}$}%
}}}}
\put(7381,-3436){\makebox(0,0)[lb]{\smash{{\SetFigFont{12}{24.0}{\rmdefault}{\mddefault}{\updefault}{\color[rgb]{0,0,0}$(S_1S_3|S_2)$}%
}}}}
\end{picture}%
\end{center}
}{
               +--------+   {x, x&-y}   +----------+
(S1|S2S3) ---> | merger | ------------> | assesser | ---> (S1S3|S2)
               +--------+               +----------+
                   ^                         ^
                   |                         |
                   |                         |
                   +------              -----+
                             S1={x}
                          S2={y, -x&y}
                            S3={x&-y}
}
\caption{A reliability ordering to exclude}
\label{fig-non-trivial}
\end{figure}

This observation can be used as a method for testing a candidate measure of
reliability: merging can be attempted using it, and if the result is not
coherent with the original assumption then the candidate is discarded. The
typical application is with a specific source of merging that is totally
reliable: every piece of information it provides is sure, albeit possibly
incomplete; another source providing information that contradicts it cannot be
considered reliable. Still better, the result of merging can be compared with
the other sources to assess their reliability. This allows for excluding some
reliability orderings even in lack of a totally reliable source. This is shown
by Theorem~\ref{non-trivial}, which can be depicted as in
Figure~\ref{fig-non-trivial}: assuming that $S_1$ is strictly more reliable
than $S_2$ and $S_3$, one possible result of merging is $\{x, x \wedge \neg
y\}$; since both $S_1$ and $S_3$ are consistent with this set, they are equally
reliable (and more so than $S_2$), contrary to the assumption. The initial
reliability ordering can therefore be excluded.

The principle is a fixpoint of the reliability ordering of the sources: if an
ordering produces a result that is not coherent with it, it is discarded. An
alternative that is considered is that of starting on the assumption of all
sources with the same reliability and checking which reliability ordering are
obtained after merging.

Some variants are considered (bipartitions with singleton or non-singleton
sources, tripartitions, partitions on bounds and weighted merge). The next
section provides a roadmap of the results obtained in each case.

\section{Summary}
\label{summary}

Section~\ref{sources-maxsets} introduces the formalization of the setting:
sources are sets of formulae, their relative reliability is formalized as an
ordered partitions of these sets. The maxsets of a set and of an ordered
partition are defined. Some general results about maxsets follow.

Section~\ref{one} contains results about the limit case where every source
provides exactly one formula, and sources are partitioned in two classes only:
reliable or not. The following results are obtained:

\begin{itemize}

\item not all partitions are stable;

\item every set of sources has at least one stable partition;

\item the maxsets of the stable partitions are the plain maxsets of the set of
all formulae.

\end{itemize}

The last result can be seen as confirmatory: a formula should be more or less
taken into consideration depending on how truthful the others provided by the
same source are; if every source provides a single formula these formulae
should be equally believed, as plain maxsets do.

Section~\ref{many} has the simplest case of interest: sources are still
partitioned into two classes depending on whether they are reliable or not, but
they are not limited to provide a single formula each. A partition may have
multiple maxsets, some inducing the partition itself and some others doing not.
Such a partition is weakly stable, as opposed to the case of strong stability
where every maxset induces the partition itself. The results obtained for this
setting are:

\begin{itemize}

\item some weakly stable partitions are not strongly stable;

\item the maxsets of the weakly stable partitions are the plain maxsets of the
set comprising all formulae provided by all sources;

\item a weakly stable partition may lead via a maxset to another partition;
that partition has a strictly larger first class;

\item the maximal weakly stable partitions w.r.t.\ containment of their first
class are the strongly stable partitions;

\item the maxsets of the strongly stable partitions are equivalent to the
maxsets of a certain partition.

\end{itemize}

Partitions are related by maxsets: the maxset of a partition induces a possibly
different partition. This relation allows for an equivalent definition of
strong stability. It is also used in the following section, where it allows for
a definition of mild stability when strong stability is unattainable.

Section~\ref{unreliable} still maintains sources providing multiple formulae,
but partitions them in three classes rather than two: totally reliable, partly
reliable and unreliable. These divisions are called tripartitions to
distinguish them from the bipartitions of the previous sections; this is
necessary because some results relate bipartitions and tripartitions. The
following results are obtained:

\begin{itemize}

\item a plain maxset is also a maxset of its induced tripartition;

\item weakly stable bipartitions correspond to weakly stable tripartitions;

\item this is not the case for strong stability;

\item the property of containment of the first class of related bipartitions
does not extend to tripartitions;

\item rather, cycles of related tripartitions are possible;

\item some sources do not have strongly stable tripartitions;

\item a third kind of stability can be defined as maximality according to the
relation among partitions.

\end{itemize}

Maximal tripartitions may not be strongly stable, as opposed to the case of
bipartitions. Defining mild stability as maximality according to the relation
between tripartitions, every source has at least a mildly stable partition,
even if it has no strongly stable partition.

Section~\ref{general} sticks to the principle of assuming a reliability
ordering and then matching it with the result of merging, but extends the
analysis in several directions:

\begin{itemize}

\item divide the sources according to the percentage of correct formulae they
provide;

\item use an arbitrary order of the sources for expressing reliability;

\item merge by weights.

\end{itemize}

Section~\ref{directions} explores future directions. One looks particularly
applicable in practice: one or more sources are known for certain to be
completely reliable. More generally, the reliability ordering may be partially
known. Other cases considered in this section are that of unreliable sources
considered as intentionally providing false information and non-fixed
percentage bounds.

\section{Sources and maxsets}
\label{sources-maxsets}

In this article, a source is identified with the information it provides: a
source is a set of formulae. Each such set is assumed consistent. The
reliability of the sources is given as an ordered partition of the sources; no
other information is assumed known on them.

Every source is assumed to always provide information at the same level of
reliability; the case in which a ``physical'' source may for example produce
formulae of two different degrees of reliability may be captured by splitting
it in two separate sources.

The example in Figure~\ref{multiple-sources} has three sources. The first
provides formulae $kb_1$, $kb_3$ and $kb_5$, and is therefore formalized by the
set $S_1=\{kb_1, kb_3, kb_5\}$. The same for the second source $S_2=\{kb_4,
kb_7\}$ and the third $S_3=\{kb_2, kb_6, kb_8\}$.

The relative reliability of the sources is represented by an ordered partition
over the set of sources. Given the three sources $S_1$, $S_2$ and $S_3$, the
situation in which the second is more reliable than the other two, which
compare the same, is represented by $(S_2 | S_1 S_3)$. More generally, such an
ordered partition is denoted by a sequence separated by the symbol $|$, where
each part is a set of sources; the most reliable ones appear before the first
$|$, the other follows in order of decreasing reliability.

While $(S_2 | S_1 S_3)$ denotes three sources, the similar partition $(S_2 |
S_1 \cup S_3)$ represents only two. Yet, the formulae and their reliability are
the same: once sources are framed into an ordered partition, merging can be
done by first flattening each class into a single set of formulae. This results
in an ordered partition of formulae, similar to that used in iterated belief
revision~\cite{rott-93}, which is in fact the same as belief merging from
sources of unequal reliability~\cite{delg-dubo-lang-06}.

Ordered partitions of formulae allow for the use of maxsets. The most common
definition is from a single-class partition $(P)$ or, equivalently, the set of
formulae $P$:

\begin{definition}
\label{plain-maxset}

A maxset of a set of formulae $P$ is a consistent subset of $P$ such that no
other consistent subset of $P$ strictly contains it.

\end{definition}

The word ``maxset'' is an abbreviation for {\em max}imal consistent sub{\em
set}. If $P$ is consistent then its only maxset is $P$ itself. Otherwise, $P$
can be made consistent by removing some of its formulae. This can be done in
various ways, leading to different maxsets. For example, $\{x, \neg y, y\}$ can
be turned consistent by removing either $\neg y$ or $y$, leading to the two
maxsets $\{x, y\}$ and $\{x, \neg y\}$. Maximality ensures that not too much is
removed: for example, $\{x\}$ is also a consistent subset, but deleting both
$y$ and $\neg y$ for restoring consistency is an overkill.

The maxsets of a single-class partition $(P)$ are the maxsets of the set $P$. A
multiple-class partition $(P_1|P_2|\ldots|P_m)$ represents an ordering of
preference over formulae in which $P_1$ ranks first, $P_2$ second, and so on.
If $P_1 \cup P_2 \cup \cdots \cup P_m$ is consistent this ordering does not
matter, otherwise it is used to decide which formulae to remove to restore
consistency. For example, $(\neg y| x, y)$ means that $\neg y$ is to be
preferred over both $x$ and $y$; therefore, the best way to restore consistency
is to remove $y$, obtaining $\{\neg y, x\}$; the second formula $x$ is
retained, even if it was in the second class of the partition, because it was
not involved in the contradiction; in other words, its removal would not help
restoring satisfiability. The principle is: remove formulae only if it is
necessary to make the set consistent, starting from the first
class~\cite{nebe-92,nebe-98,rott-93,delg-dubo-lang-06}.

\begin{definition}
\label{maxset}

A maxset of a partition $(P_1|\ldots|P_m)$ is a subset $M \subseteq P_1 \cup
\cdots \cup P_m$ such that $M \cap (P_1 \cup \cdots \cup P_i)$ a maxset of
$(P_1|\cdots|P_i)$ for all $1 \leq i \leq m$.

\end{definition}

By virtue of the universal quantifier over $i$, $M \cap P_1$ needs to be a
maxset of $P_1$. Only when this condition is met the check with $P_2$ matters.
This ensures that formulae of the first class are removed only if this is
really necessary to remove contradiction, even disregarding all other formulae.
This way, maximality within $P_1$ takes precedence over maximality within
$P_2$, which then take precedence over maximality within $P_3$, etc.

Collapsing $(P_1|\ldots|P_m)$ into $P_1 \cup \cdots \cup P_m$ makes all
formulae compare the same. The maxsets of this set are called the plain maxsets
of the partition. Every maxset is also a plain maxset, but not the other way
around. Indeed, plain maxsets are obtained by first deleting the ordering over
formulae; therefore, some plain maxsets may remove formulae that occur early in
the partition and keep others that occur late.

Slightly abusing notation, $(S_1|S_2 S_3)$ means $(S_1|S_2 \cup S_3)$. More
generally, if two or more sets occur as a class of a partition, that means
their union.

\

Some results carry from maxsets of sets to maxsets of partitions. The first is
an almost direct consequence of being a maximal subset: it contains every
formula consistent with it.

\begin{lemma}
\label{consistent-entail}

If $M$ is a maxset of $(P_1|\ldots|P_m)$ and $M$ is consistent with $kb \in P_1
\cup \cdots \cup P_m$, then $M$ contains $kb$.

\end{lemma}

\proof By definition, $M \cap (P_1 \cup \cdots \cup P_i)$ is a maxset of $P_1
\cup \cdots \cup P_i$ for every $i \in \{1,\ldots,m\}$; for $i=m$ this
condition is that $M$ is a maxset of $P_1 \cup \cdots \cup P_m$. If $kb \not\in
M$ but $M \cup \{kb\}$ is consistent, then $M$ is not a maximal consistent
subset of $P_1 \cup \cdots \cup P_m$, and this condition is part of the
definition of $M$ being a maxset of $(P_1|\ldots|P_m)$.~\qed

A maxset of a partition is also a maxset of the set comprising all formulae in
the partition. The converse does not hold in general. For example, $(x|\neg x)$
has a single maxset $\{x\}$, but the set $\{x,\neg x\}$ has the two maxsets
$\{x\}$ and $\{\neg x\}$. The second is not a maxset of the partition because
it prefers the lower-ranked formula $\neg x$ over the higher-ranked $x$.

If a set it consistent its only maxset is itself. A similar result holds for
partitions in different ways: for $P_1$ only, for $P_1 \cup \cdots \cup P_m$,
or for every union in between.

\begin{lemma}
\label{consistent-highest}

If $P_1 \cup \cdots \cup P_i$ is consistent, every maxset of $(P_1|\ldots|P_m)$
is a superset of $P_1 \cup \cdots \cup P_i$.

\end{lemma}

\proof By definition of maxsets of a partition, $M \cap (P_1 \cup \cdots \cup
P_i)$ is a maxset of $P_1 \cup \cdots \cup P_i$. Since this union is
consistent, its only maxset is itself:
{} $M \cap (P_1 \cup \cdots \cup P_i) = P_1 \cup \cdots \cup P_i$.
This implies that $M$ contains all of $P_1 \cup \cdots \cup P_i$~\qed

The converse holds for partitions made of two classes.

\begin{lemma}
\label{plain-allp}

If a maxset of $P_1 \cup P_2$ is a superset of $P_1$, it is a maxset of 
$(P_1|P_2)$.

\end{lemma}

\proof Let $M$ be a maxset of $P_1 \cup P_2$ such that $P_1 \subseteq M$. By
definition, $M$ is a maxset of $(P_1|P_2)$ if it is a maxset of $P_1 \cup P_2$
and $M \cap P_1$ is a maxset of $P_1$. The first condition holds by assumption.
The second holds as well: $M$ being a maxset, it is consistent; therefore, its
subset $M \cap P_1$ is consistent; it is maximally consistent within $P_1$
because $M \cap P_1 = P_1$.~\qed

Given a partition $(P_1|\ldots|P_m)$, its plain maxsets are obtained by
flattening the partition into a set $P_1 \cup \cdots \cup P_m$. A maxset is
also a plain maxset; the converse may hold or not. It holds for two-class
partitions under a simple condition.

\begin{lemma}
\label{first-class}

A plain maxset $M$ is a maxset of a bipartition $(P_1|P_2)$ if no other plain
maxset $M'$ is such that $M \cap P_1 \subset M' \cap P_1$.

\end{lemma}

\proof For two-classes partitions, the definition of maxsets specializes to: $M
\cap P_1$ is a maxset of $P_1$ and $M$ is a maxset of $P_1 \cup P_2$. The
second part is the same as the definition of a plain maxset. The first part is
ensured by the statement of the lemma.~\qed

The case of partitions of two classes is the simplest considered in this
article: sources are assessed as either ``reliable'' or ``unreliable''. For
example, if the source providing $S_1$ and $S_3$ are reliable while $S_2$ is
unreliable, the partition is $(S_1 S_3|S_2)$. With the abuse of notation
announced above, this is the same as $(S_1 \cup S_3|S_2)$.

Counterexamples are based on formulae having a given set of maxsets. The
following lemma allows avoiding building them explicitly and instead just write
for example ``formulae $A$, $B$ and $C$ such that their maxsets are $\{A,B\}$
and $\{B,C\}$''.

\begin{lemma}[\cite{libe-16}]
\label{synthesis}
\nop

Given some sets of letters, none contained in another, there exists a formula
for each letter so that the maxsets of these formulae correspond to the given
sets of letters.

\end{lemma}

\section{Two-class partitions of singleton sources}
\label{one}

Sources can be either reliable or unreliable. As a further simplification, each
may provide a single formula. Both assumptions are released in the following
sections; yet, this very simple case already allows for some significant
results.

If the classification of sources is given, each maxset of the ordered partition
is a way for merging them while retaining as much of them as consistency
allows. If the sources
{} $S_1=\{x\}$,
{} $S_2=\{\neg x \wedge \neg y\}$ and
{} $S_3=\{\neg x\}$
are partitioned into $(S_1 S_3|S_2)$, their maxsets are
{} $\{x\}$ and
{} $\{\neg x, \neg x \wedge \neg y\}$.
As it is common in belief revision (and in nonmonotonic reasoning in general),
when multiple results are possible and nothing suggests that one should be
preferred over the others, they are disjoined. In this case, the disjunction of
the maxsets reflects the impossibility of choosing one over the other.

When the reliability of the source is not given, the same procedure can be
carried on the one-class partition $(S_1 S_2 S_3)$, leading to all plain
maxsets of $S_1 \cup S_2 \cup S_3$. Alternatively, one may arbitrarily assume a
different partition, like $(S_1 S_3|S_2)$, and then check whether the result of
merging is consistent with the assumption. In this case, the maxsets are
$\{x\}$ and $\{\neg x, \neg x \wedge \neg y\}$. The first is consistent with
$S_1$: therefore, the assumption that $S_1$ is reliable is not contradicted.
The same holds for $S_3$. However, $S_2$ also is consistent with the second
maxset, contrary to its assumed unreliability.

The following sections use the same principle, which in this simple case can be
summarized as:

\begin{itemize}

\item assume an arbitrary partition of the sources (reliable and unreliable);

\item determine a way to merge them (a maxsets);

\item use it to classify the sources: the ones consistent with it are reliable,
the others are unreliable;

\item check if this classification is the same as that initially assumed.

\end{itemize}

This section assumes that sources are singletons (each source provides exactly
one formula) and that partitions have two classes (reliable and unreliable).
The following sections release both constraints, but the general principle
stays the same: assume some relative reliability of the sources, merge, use the
result for assessing the reliability of the sources; if this is not the same as
that originally assumed, it is discarded.

Assuming a partition of the sources, each maxset is a way to merge them. What
is missing is the third step of the process: from the result of merging, assess
the reliability of the sources. This is done by checking the consistency of the
sources with each maxset $M$.

\begin{definition}

A set $M$ induces on the set of sources $S$ the bipartition $(R|S \backslash
R)$, where $R$ contains the sources that are consistent with $M$.

\end{definition}

This definition closes the circle: from a partition one can determine its
maxsets, and each maxset induces a partition of the source. Technicalities
apart, assuming a reliability ordering of the sources allows merging; its
result allows assessing the reliability of the sources.

\begin{definition}

A partition is {\em stable} if it has a maxset whose induced partition is the
partition itself.

\end{definition}

Stability could also be defined by requiring all maxsets to induce the
partition itself. This stronger condition is used in the next sections. For
this introductory part of the article, the weaker definition is used: if there
is a possible way to perform the merge (a maxset) such that the result is
consistent with a source, the source is considered reliable. The results
obtained from this definition are:

\begin{itemize}

\item stable partitions always exist;

\item not all partitions are stable;

\item the maxsets of the stable partitions are the plain maxsets.

\end{itemize}

The first two facts show that the definition of stability is not trivial: some
reliability orderings lead to merging results that contradict them and some
others do not; the constraint of stability gives some useful information in
removing some but not all of them from consideration. The third fact may look
disappointing: even with the condition of stability still all plain maxsets are
obtained. In the next sections the assumption of singleton sources and binary
partitions will be lifted, and this result no longer holds. For now, it can be
seen as a confirmation that in the basic case the mechanism is sensible:
lacking other formulae coming from the same source, the reliability of a
formula depends only on itself; therefore, nothing allows assuming a formula
more reliable than another.

\begin{theorem}
\label{non-trivial}

There exist three formulae $A$, $B$ and $C$ such that the partition $(AB|C)$ is
stable while $(A|BC)$ is not.

\end{theorem}

\proof Three consistent formulae $A$, $B$ and $C$ are chosen so that their
plain maxsets are $\{A,B\}$ and $\{B,C\}$. Lemma~\ref{synthesis} proves that
this is possible because these sets are not contained in each other; for
example, $A=x \wedge y$, $B=x$ and $C=x \wedge \neg y$.

Formulae with such maxsets can be graphically depicted in the space of models,
where each formula is a box representing its set of models and intersections
indicate mutual consistency.

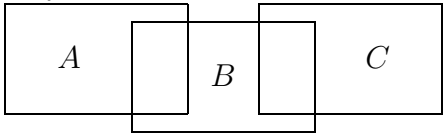
\setlength{\unitlength}{5000sp}%
\begingroup\makeatletter\ifx\SetFigFont\undefined%
\gdef\SetFigFont#1#2#3#4#5{%
  \reset@font\fontsize{#1}{#2pt}%
  \fontfamily{#3}\fontseries{#4}\fontshape{#5}%
  \selectfont}%
\fi\endgroup%
\begin{picture}(2184,654)(4399,-3313)
\thinlines
{\color[rgb]{0,0,0}\put(4411,-3211){\framebox(900,540){}}
}%
{\color[rgb]{0,0,0}\put(5041,-3301){\framebox(900,540){}}
}%
{\color[rgb]{0,0,0}\put(5671,-3211){\framebox(900,540){}}
}%
\put(4726,-2986){\makebox(0,0)[b]{\smash{{\SetFigFont{12}{24.0}{\rmdefault}{\mddefault}{\itdefault}{\color[rgb]{0,0,0}$A$}%
}}}}
\put(5491,-3076){\makebox(0,0)[b]{\smash{{\SetFigFont{12}{24.0}{\rmdefault}{\mddefault}{\itdefault}{\color[rgb]{0,0,0}$B$}%
}}}}
\put(6256,-2986){\makebox(0,0)[b]{\smash{{\SetFigFont{12}{24.0}{\rmdefault}{\mddefault}{\itdefault}{\color[rgb]{0,0,0}$C$}%
}}}}
\end{picture}%
\nop{
+-------+     +-------+
|  A  +-+-----+-+  C  |
|     | |  B  | |     |
+-----+-+     +-+-----+
      +---------+
}

The partition $(AB|C)$ is stable. Since $\{A,B\}$ is consistent but conflicts
with $C$, it is the only maxset of the partition. The reliability of the
sources can be assessed from it. Since $A$ and $B$ are consistent with
$\{A,B\}$, their sources $\{A\}$ and $\{B\}$ are in the first class of the
partition induced by the maxset. Instead, $C$ is inconsistent with the maxset.
Therefore, its source $\{C\}$ is in the second class of the induced partition.
The resulting partition is therefore $(AB|C)$, the same assumed in the first
place.

The partition $(A|BC)$ is not stable. Indeed, the plain maxset $\{B,C\}$ is
excluded because it does not contain the most reliable formula $A$. The only
plain maxset that is also a maxset of the partition is $\{A,B\}$. This set is
consistent with $A$ and $B$ but not with $C$. Therefore, the induced partition
is $(AB|C)$, while that initially assumed was $(A|BC)$.~\qed

The maxsets of $\{A,B,C\}$ are $\{A,B\}$ and $\{B,C\}$. The first is generated
by the stable partition $(AB|C)$; by symmetry, the second is generated by
$(BC|A)$. In this particular case, both maxsets are generated by some stable
partition. More generally, this happens whenever all sources contain a single
formula and partitions are made of two classes.

\begin{lemma}

If $S$ is a set of singleton sources, $R \subseteq S$ and $(R|S \backslash R)$
is a stable partition, then $R$ is its only maxset.

\end{lemma}

\proof By definition of stability, $(R|S \backslash R)$ has a maxset $M$ such
that $R$ is exactly the set of elements of $S$ that are consistent with $M$.
Since every $kb \in R$ is consistent with $M$, Lemma~\ref{consistent-entail}
implies $kb \in M$. In other words, $R \subseteq M$.

Since $R$ is the set of elements of $S$ that are consistent with $M$, all other
elements of $S$ are inconsistent with $M$, and are therefore not in $M$. This
implies that $M=R$.

Let $M'$ be an arbitrary maxset of $(R|S \backslash R)$. Since $R=M$ and $M$ is
consistent by definition of maxset, so is $R$. By
Lemma~\ref{consistent-highest}, every maxset of $(R|S \backslash R)$ is a
superset of $R$, that is, $R \subseteq M'$. But no element of $S \backslash R$
is consistent with $M=R$, implying that none is consistent with $M'$. This
proves that $M'=M$.~\qed

This lemma is used to prove that the stable partitions generate exactly the
plain maxsets of the set of sources if sources are singletons and partitions
comprise two classes.

\begin{theorem}

$R \subseteq S$ is a plain maxset of the set of singleton sources $S$ if and
only if $(R|S \backslash R)$ is stable.

\end{theorem}

\proof By the above lemma, if $(R|S \backslash R)$ is stable then $R$ is its
only maxset. By definition of maxsets, it is also a maxset of $R \cup (S
\backslash R) = S$.

Let $R$ be a plain maxset of $S$. For it being a maxset of a partition
$(P_1|P_2)$ it has to be a maxset of $P_1 \cup P_2$ and $R \cap P_1$ be a
maxset of $P_1$. In this particular case, $P_1 = R$ and $P_2 = S \backslash R$.

\begin{itemize}

\item The first condition holds because $P_1 \cup P_2 = R \cup (S \backslash R)
= S$, and $R$ is a maxset of $S$ by assumption.

\item The second condition holds because $R \cap P_1 = R \cap R = R$. This is a
maxset of $P_1$ because $R$ is consistent (being also a maxset of $S$) and
because no other element of $P_1$ exist at all.~\qed

\end{itemize}

The maxsets of the stable partitions can be taken as the result of merging. The
above theorem proves that in the case of singleton sources and two-class
partitions, this is the same as selecting the plain maxsets of the set of
formulae provided by the sources. It also proves that selecting only the
maxsets that lead to the same partitions lead to the same result. This was
otherwise a possible other definition of the result of merging: instead of
selecting all maxsets of all stable partitions, select a maxsets $M$ only the
induced partition $(R|S \backslash R)$ is stable and in turn generates $M$.
However, the theorem proves that all maxsets would be obtained.

\section{Two-class partitions of non-singleton sources}
\label{many}

The final theorem of the previous section shows that no selection stems from
assuming and then checking the reliability of single-formulae sources: all
plain maxsets derive from the stable partitions. This is unsurprising, since
the reliability of a formula is the same as the reliability of its source, and
could only be altered by the other formulae in its source. This is the case
considered in this section: sources with multiple formulae.

\begin{definition}

A set of formulae $M$ induces on a set of sources $S$ the partition $(R|S
\backslash R)$ where $R$ is the set of sources whose formulae are all
consistent with $M$.

\end{definition}

The partition $(R|S \backslash R)$ classes the sources as either reliable
(those in $R$, whose formulae are all consistent with $M$) and unreliable (the
others). In this section sources are considered either reliable or not, but
sources may also be considered partly reliable. This three-class division is
analyzed in the next section. Other variants are discussed at the end of the
article.

The partition is on the sources, not on formulae. As a result, even if the same
formula is provided by two sources, the first may be in $R$ but the second does
not.

Reliability is assumed and then checked with a fixpoint definition.

\begin{definition}

An ordered partition $(R|S \backslash R)$ of the sources is {\bf weakly stable}
if it has a maxset $M$ that induces the partition itself.

\end{definition}

This definition requires only that one of the maxsets of the partition to
induce the partition itself. But a partition may have multiple maxsets; the
others may induce different partitions. Requiring the condition to hold for all
of them will be shown to have different consequences.

\begin{definition}

An ordered partition $(R|S \backslash R)$ of the sources is {\bf strongly
stable} if all its maxsets induce the partition itself.

\end{definition}

A different way to restrict the maxsets is to still consider all weak
partitions, but only their maxsets that induce the partition itself. It will
however been shown that the other maxsets induce weakly stable partitions. In
other words, the maxsets that do not induce the partition itself do that for
some other weak partition.

Yet another restriction is to start from the partition having all sources in
its first class. This realizes the assumption that all sources are initially
compared the same. Following the maxsets and the induced partitions may then
lead to some weakly or strongly stable partitions. This may seem a restriction
since only partitions reachable from the all-equal one are considered. As shown
below, it is not.

The results proved in this section for non-singleton sources and
reliable--unreliable partitions are:

\begin{itemize}

\item restricting to partitions reachable from the all-equal one does not
eliminate any weakly or stable partition;

\item some weakly stable partitions are not strongly stable;

\item a weakly or strongly stable partition may have more than one maxset;

\item the maxsets of the weakly stable partitions are the plain maxsets;

\item a weakly stable partition may have a maxset that induces a different
partition; however, the first class of the latter is larger than that of the
former;

\item the maximal weakly stable partition w.r.t.\  containment of their first
classes are the strongly stable partitions;

\item the maxsets of the strongly stable partitions are the maxsets of the
partition $(\wedge S_1 \ldots \wedge S_n|S_1 \cup \cdots \cup S_n)$.

\end{itemize}

Considering only partitions reachable from the partition that has all sources
in its first class is not a restriction at all, even if reachability is limited
to a single step.

\begin{theorem}
\label{reachable}

If a partition is weakly or strongly stable then it is induced by a maxset that
is also a maxset of the partition where all sources are in the first class.

\end{theorem}

\proof Since the partition is stable it has a maxset $M$ that induces itself.
By definition, $M$ is also a plain maxset. The partition where all sources are
in the first class has exactly all plain maxsets, including $M$.~\qed

Every strongly stable partition is also weakly stable, as an immediate
consequence of the definition. The converse does not hold: a weakly stable
partition may have a maxset that induces a partition that is different from the
original one.

\begin{theorem}
\label{not-strong}

There exists a set of sources whose weakly stable partitions are not all
strongly stable.

\end{theorem}

\proof The sources are $\{A\}$, $\{B,C\}$ and $\{D\}$, the plain maxsets are
$\{A,B\}$, $\{B,C\}$, $\{A,D\}$. This is a correct description of a merging
scenario because sources are identified with the set of formulae they provide,
and formulae can be specified by giving their maxsets thanks to
Lemma~\ref{synthesis}. As an example,
{} $A=x$,
{} $B=y$,
{} $C=\neg x \wedge y$ and
{} $D=x \wedge \neg y$.
The situation can be depicted in the space of models as follows.

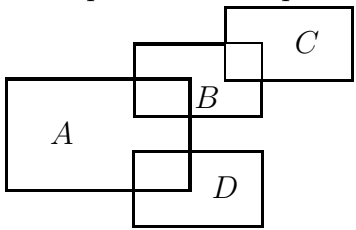
\setlength{\unitlength}{5000sp}%
\begingroup\makeatletter\ifx\SetFigFont\undefined%
\gdef\SetFigFont#1#2#3#4#5{%
  \reset@font\fontsize{#1}{#2pt}%
  \fontfamily{#3}\fontseries{#4}\fontshape{#5}%
  \selectfont}%
\fi\endgroup%
\begin{picture}(1754,1124)(4389,-3503)
\thinlines
{\color[rgb]{0,0,0}\put(5041,-2941){\framebox(630,360){}}
}%
{\color[rgb]{0,0,0}\put(5491,-2761){\framebox(630,360){}}
}%
\thicklines
{\color[rgb]{0,0,0}\put(4411,-3301){\framebox(900,540){}}
}%
{\color[rgb]{0,0,0}\put(5041,-3481){\framebox(630,360){}}
}%
{\color[rgb]{0,0,0}\put(5491,-2401){\line( 0,-1){180}}
\put(5491,-2581){\line(-1, 0){450}}
\put(5041,-2581){\line( 0,-1){360}}
\put(5041,-2941){\line( 1, 0){630}}
\put(5671,-2941){\line( 0, 1){180}}
\put(5671,-2761){\line( 1, 0){450}}
\put(6121,-2761){\line( 0, 1){360}}
\put(6121,-2401){\line(-1, 0){630}}
}%
\put(4681,-3076){\makebox(0,0)[b]{\smash{{\SetFigFont{12}{24.0}{\rmdefault}{\mddefault}{\itdefault}{\color[rgb]{0,0,0}$A$}%
}}}}
\put(5401,-2896){\makebox(0,0)[b]{\smash{{\SetFigFont{12}{24.0}{\rmdefault}{\mddefault}{\itdefault}{\color[rgb]{0,0,0}$B$}%
}}}}
\put(5896,-2626){\makebox(0,0)[b]{\smash{{\SetFigFont{12}{24.0}{\rmdefault}{\mddefault}{\itdefault}{\color[rgb]{0,0,0}$C$}%
}}}}
\put(5491,-3346){\makebox(0,0)[b]{\smash{{\SetFigFont{12}{24.0}{\rmdefault}{\mddefault}{\itdefault}{\color[rgb]{0,0,0}$D$}%
}}}}
\end{picture}%
\nop{
             +--------+
             |     C  |
       +-----+...     |
       | B   .  .     |
       |     ...+-----+
+------+---+    |
|      +---+----+
|  A       |
|      +---+----+
+------|---+    |
       |        |
       |     D  |
       +--------+
} 

All maxsets of the partition $(\{A\}|\{B,C\}\{D\})$ contain $A$ by
Lemma~\ref{consistent-highest}. Of the formulae of the second class, only $B$
and $D$ are consistent with $A$, leading to the maxsets $\{A,B\}$ and
$\{A,D\}$. Their induced partitions are:

\begin{description}

\item{$\{A,B\}$:} the only source that contains only formulae consistent with
this maxset is $\{A\}$; as a result, the induced partition is
$(\{A\}|\{B,C\}\{D\})$ itself, showing that this partition is weakly stable;

\item{$\{A,D\}$:} this maxset is consistent with the two sources $\{A\}$ and
$\{D\}$; the induced partition is therefore $(\{A\}\{D\}|\{B,C\})$, which is
not the same as the partition assumed initially.

\end{description}

The first point shows that $\{A\}|\{B,C\}\{D\})$ is weakly stable, since it has
a maxset that induces the partition itself. The second shows that it is not
strongly stable, since the other maxset induces a different partition.~\qed

A weakly stable partition having a single maxset is strongly stable by
definition. However, some strongly stable partitions have more than one maxset,
as the following example shows.

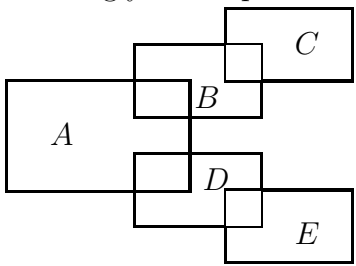
\setlength{\unitlength}{5000sp}%
\begingroup\makeatletter\ifx\SetFigFont\undefined%
\gdef\SetFigFont#1#2#3#4#5{%
  \reset@font\fontsize{#1}{#2pt}%
  \fontfamily{#3}\fontseries{#4}\fontshape{#5}%
  \selectfont}%
\fi\endgroup%
\begin{picture}(1754,1304)(4389,-3683)
\thinlines
{\color[rgb]{0,0,0}\put(5041,-2941){\framebox(630,360){}}
}%
{\color[rgb]{0,0,0}\put(5491,-2761){\framebox(630,360){}}
}%
\thicklines
{\color[rgb]{0,0,0}\put(4411,-3301){\framebox(900,540){}}
}%
{\color[rgb]{0,0,0}\put(5491,-2401){\line( 0,-1){180}}
\put(5491,-2581){\line(-1, 0){450}}
\put(5041,-2581){\line( 0,-1){360}}
\put(5041,-2941){\line( 1, 0){630}}
\put(5671,-2941){\line( 0, 1){180}}
\put(5671,-2761){\line( 1, 0){450}}
\put(6121,-2761){\line( 0, 1){360}}
\put(6121,-2401){\line(-1, 0){630}}
}%
\thinlines
{\color[rgb]{0,0,0}\put(5041,-3481){\framebox(630,360){}}
}%
{\color[rgb]{0,0,0}\put(5491,-3661){\framebox(630,360){}}
}%
\thicklines
{\color[rgb]{0,0,0}\put(5671,-3121){\line( 0,-1){180}}
\put(5671,-3301){\line( 1, 0){450}}
\put(6121,-3301){\line( 0,-1){360}}
\put(6121,-3661){\line(-1, 0){630}}
\put(5491,-3661){\line( 0, 1){180}}
\put(5491,-3481){\line(-1, 0){450}}
\put(5041,-3481){\line( 0, 1){360}}
\put(5041,-3121){\line( 1, 0){630}}
}%
\put(4681,-3076){\makebox(0,0)[b]{\smash{{\SetFigFont{12}{24.0}{\rmdefault}{\mddefault}{\itdefault}{\color[rgb]{0,0,0}$A$}%
}}}}
\put(5401,-2896){\makebox(0,0)[b]{\smash{{\SetFigFont{12}{24.0}{\rmdefault}{\mddefault}{\itdefault}{\color[rgb]{0,0,0}$B$}%
}}}}
\put(5896,-2626){\makebox(0,0)[b]{\smash{{\SetFigFont{12}{24.0}{\rmdefault}{\mddefault}{\itdefault}{\color[rgb]{0,0,0}$C$}%
}}}}
\put(5446,-3301){\makebox(0,0)[b]{\smash{{\SetFigFont{12}{24.0}{\rmdefault}{\mddefault}{\itdefault}{\color[rgb]{0,0,0}$D$}%
}}}}
\put(5896,-3571){\makebox(0,0)[b]{\smash{{\SetFigFont{12}{24.0}{\rmdefault}{\mddefault}{\itdefault}{\color[rgb]{0,0,0}$E$}%
}}}}
\end{picture}%
\nop{
             +--------+
             |     C  |
       +-----+...     |
       | B   .  .     |
       |     ...+-----+
+------+---+    |
|      +---+----+
|  A       |
|      +---+----+
+------|---+    |
       |     ...+-----+
       | D   .  .     |
       +-----+--+     |
             |     E  |
             +--------+
}

The partition $(\{A\}|\{B,C\}\{D,E\})$ has two maxsets: $\{A,B\}$ and
$\{A,D\}$. The source $\{A\}$ is the only one having only formulae consistent
with the first maxset, leading to partition itself. The same for the second
maxset. This is a strongly stable partition having two maxsets.

As in the case of singleton sources, the maxsets of the weakly stable
partitions are exactly all the plain maxsets of the set of sources. This is
proved from a property of the plain maxsets.

\begin{lemma}
\label{itself}

If $M$ is a plain maxset, its induced partition has $M$ as a maxset.

\end{lemma}

\proof Let $(R|S \backslash R)$ be the partition induced by $M$. By definition,
all formulae in $R$ are consistent with $M$. If some of these formulae were not
in $M$, then $M$ would not be a plain maxset of $S$. This proves that $M$
contains all formulae in $R$. By Lemma~\ref{plain-allp}, $M$ is a maxset of
$(R|S \backslash R)$.~\qed

This lemma shows that starting from a plain maxset and moving to its induced
partition and then to its maxsets, among them is the original maxset. The
partition may also have other maxsets, as shown in the proof of
Theorem~\ref{not-strong}: the maxset $\{A,B\}$ induces the partition
$(\{A\}|\{B,C\}\{D\})$, which has two maxsets: $\{A,B\}$ itself and $\{A,D\}$.

This lemma has two side consequences.

The first is related to an additional constraint that was considered above:
instead of accepting all maxsets of all weakly stable partition, restrict to
those generating the partition. In other words, if a weakly stable partition
$(P_1|P_2)$ has the maxset $M$, but this maxset induces a different partition,
exclude $M$ from consideration. This restriction is useless: even if $M$
induces a different partition $(P_1'|P_2')$, one of the maxsets of this
partition is $M$ itself.

The second consequence is that every set of sources has at least a weakly
stable partition. Indeed, every set has at least a maxset $M$, and this maxset
induces a partition that has $M$ has a maxset. That partition is therefore
weakly stable.

The converse of the lemma holds by definition: the maxsets of a partition are
some of the plain maxsets.

\begin{corollary}

The maxsets of weakly stable partitions are the plain maxsets.

\end{corollary}

This corollary tells which the maxsets of the weakly stable partitions are: the
plain maxsets. However, it is also useful in the other way around. The most
obvious way for finding a plain maxset is to start from the empty set and then
iteratively adding a formula if consistent with the set. When formulae are
provided in a set of sources, this method can be slightly simplified because
$M$ has to be consistent with all formulae in the first class of some
partition.

\begin{algorithm}[Maxset and induced partition]

\

\begin{enumerate}

\item $M = \emptyset$

\item $R=\emptyset$

\item for each source $S_i \in S$:

\begin{enumerate}

\item[] if $M \cup S_i$ is consistent:

\begin{enumerate}

\item $M = M \cup S_i$

\item $R = R \cup \{S_i\}$

\end{enumerate}

\end{enumerate}

\item output the weakly stable partition $(R | S \backslash R)$

\item for each $S_i \in S \backslash R$

\begin{enumerate}

\item[] for each $F \in S_i$

\begin{enumerate}

\item[] if $S_i \backslash \{F\} \not\subseteq M$
and $M \cup \{F\}$ is consistent then

\begin{enumerate}

\item[] $M = M \cup \{F\}$

\end{enumerate}

\end{enumerate}

\end{enumerate}

\item output maxset $M$

\end{enumerate}

\end{algorithm}

The additional check $S_i \backslash \{F_i\} \subseteq M$ allows saving some
consistency checks, especially for sources of little cardinality: since $S_i$
is not in the first class of the partition induced by $M$, not all its formulae
can be in $M$. Therefore, if all of them but $F$ are in $M$, the inconsistency
of $M \cup \{F\}$ is already proved.

This algorithm is guaranteed to find a maxset if all sources are consistent
sets, as assumed in this article. However, some maxsets are not found this way,
like
{} $\{x,y\}$
for the sources
{} $\{\{x, x \wedge \neg y\},\{y, \neg x \wedge y\}\}$.
Only maxsets of partitions with a non-empty first class are obtained.

Since a partition may have a maxset $M$ whose induced partition is not itself,
the question of which other maxsets that partition has raises. The previous
theorem proves only that one of them is $M$, but tells nothing about the
others. Of particular interest are their induced partitions.

\begin{theorem}
\label{allp}

If $(R|S \backslash R)$ is weakly stable, then all its maxsets include $\cup R$
and induce a partition $(R'|S \backslash R')$ such that $R \subseteq R'$.

\end{theorem}

\proof Since the partition is weakly stable it has a maxset $M$ that induces
the same partition $(R|S \backslash R)$. By definition of induced partition,
all formulae in $R$ are consistent with $M$. By Lemma~\ref{consistent-entail},
since $M$ is a maxset then it contains them as well. This proves that $R$ is
contained in $M$; therefore, $R$ is consistent; by
Lemma~\ref{consistent-highest}, all maxsets of $(R|S \backslash R)$ include it
and all their induced partitions have all of $R$ in their first class.~\qed

A partition $(R|S \backslash R)$ may have a maxset that induces $(R'|S
\backslash R')$ with $R' \not= R$, but this implies $R \subset R'$. When moving
from partition to maxset and then to another partition, either the partitions
are the same or the first class is strictly increasing. This forbids the
existence of cycles partition---maxset---partition---... unless the partitions
are the same. From a different angle, since the first class is strictly
increasing, at some point a maximum is hit.

\begin{theorem}
\label{maximal}

The maximal weakly stable partitions w.r.t.\  containment of their first class
are the strongly stable partition.

\end{theorem}

\proof Let $(R|S \backslash R)$ be a weakly stable partition that is maximal on
set-containment of its first class. It is not strongly stable only if one of
its maxsets $M$ induces a partition $(R'|S \backslash R')$ with $R \not= R'$.
By Lemma~\ref{allp}, $R \subset R'$. Since $(R'|S \backslash R')$ is induced by
$M$, it has $M$ has a maxset by Lemma~\ref{itself}. Therefore, this partition
is weakly stable and has its first class strictly containing $R$, contrary to
the maximality of $(R|S \backslash R)$.

The converse is proved by assuming that $(R|S \backslash R)$ is strongly
stable, and proving that it is a maximal weakly stable partition. That it is
weakly stable is an immediate consequence of the definitions of stability.
Remain to prove that it is maximal. Let $(R'|S \backslash R')$ be another
weakly stable partition with $R \subset R'$. By definition of weak stability,
$(R'|S \backslash R')$ has a maxset $M$ whose induced partition is $(R'|S
\backslash R')$. By Theorem~\ref{allp}, $M$ contains $R'$, which in turn
contains $R$. Since every maxset is also a plain maxset, and $M$ contains $R$,
by Lemma~\ref{plain-allp} $M$ is a maxset of $(R|S \backslash R)$. As a result,
$M$ is a maxset of $(R|S \backslash R)$ but its induced partition is $(R'|S
\backslash R')$, contradicting the assumption of strong stability.~\qed

This proves the existence of strongly stable partitions for all sources, and
also establish a way to find them from weakly stable partitions: following the
maxsets and then the induced partitions.

Lemma~\ref{itself} and Theorem~\ref{maximal} provide a way for finding weakly
and strongly stable partitions.

\begin{algorithm}[Weakly and strongly stable partitions of two classes]

\

\begin{enumerate}

\item set $P$ to be an arbitrary partition, for example $(S_1 \ldots S_n|)$

\item find a maxset $M$ of $P$
\label{first}

\item determine the partition $P'$ that $M$ induces

\item output $P'$ as a weakly stable partition (Lemma~\ref{itself})
\label{output-weak}

\item for each maxset $M'$ of $P'$:
\label{step-maxset}

\begin{enumerate}

\item[] if $M'$ induces a partition $P'' \ne P'$ then:

\begin{enumerate}

\item set $P'=P''$

\item go to Step~\ref{step-maxset}

\end{enumerate}

\end{enumerate}

\item output $P'$ as strongly stable partition (Theorem~\ref{maximal})
\label{output-strong}

\end{enumerate}

\end{algorithm}

This algorithm works because every maxset $M$ of a partition $P$ is also a
plain maxset by definition, and the partition induced by $M$ has $M$ as a
maxset thanks to Lemma~\ref{itself}. Using $P=(S_1 \ldots S_n|)$ as the initial
partition allows for finding all weakly and stable partitions rather than just
one. Indeed, the maxsets of this partitions are exactly the plain maxsets;
therefore, all weakly stable partitions can be output in
Step~\ref{output-weak}, and all strongly stable partitions in
Step~\ref{output-strong} by considering all maxsets instead of just one.

The strongly stable partitions can also be found by a specific partition.

\begin{theorem}
\label{partition-of-strong}

The maxsets of $(\wedge S_1 \ldots \wedge S_n | S_1 \cup \cdots \cup S_n)$ are
exactly the maxsets of the strongly stable partitions of the sources
$S_1,\ldots,S_n$ with the addition to each of every $\wedge S_i$ consistent
with it.

\end{theorem}

\proof Let
{} $S=\{S_1,\ldots,S_n\}$ and
{} $S^\wedge=\{\wedge S_1, \ldots, \wedge S_n\}$.
The partition in the statement of the theorem is therefore $(S^\wedge|\cup S)$.
Let $a(M)$ be the set obtained by adding to $M \subseteq S$ all formulae
$\wedge S_i \in S^\wedge$ consistent with it.

\

Let $M$ be a maxset of a strongly stable partition $(R|S \backslash R)$ and
prove that $a(M)$ is a maxset of $(S^\wedge|\cup S)$.

If $M$ is consistent with $\wedge S_i$, every formula in $S_i$ is consistent
with $M$. Therefore, it is in $M$ since $M$ is also a maxset of $S$. This
implies that all formulae $\wedge S_i$ consistent with $M$ are also entailed by
$M$. Therefore, $M$ and $a(M)$ are equivalent.

Since $M$ is a maxset of $(R|S \backslash R)$, it is a maxset of $S$. Since
$a(M)$ contains all elements of $S^\wedge$ that are consistent with $M$, it is
a maxset of $S^\wedge \cup (\cup S)$. In other words, it is a plain maxset of
$(S^\wedge|\cup S)$. This is the first part of the definition of $a(M)$ being a
maxset of $(S^\wedge|\cup S)$; the other that is still to be proved is that
$a(M) \cap S^\wedge$ is a maxset of $S^\wedge$.

The converse is possible if a source is consistent with $a(M) \cap S^\wedge$
but is not contained in it. For the sake of clarity, the sources can be
re-indexed so that $a(M) \cap S^\wedge = \{\wedge S_1 \ldots \wedge S_i\}$ and
the source that is consistent but not contained in this set is $S_{i+1}$.

Using such indexes, the partition induced by $M$ is $(S_1 \ldots S_i|S_{i+1}
\ldots S_n)$; this is the same as $(R|S \backslash R)$ thanks to strong
stability. Since $M$ is a maxset, it is consistent. Therefore, its equivalent
set $a(M)$ and its subset $a(M) \cap S^\wedge$ are consistent as well. The
latter is also consistent with $\wedge S_{i+1}$ by assumption; therefore, all
formulae in $S_{i+1}$ can be added to it while retaining consistency.
Iteratively adding formulae from $S_{i+2} \ldots S_n$ produces a maxset $M'$ of
the partition $(R|S \backslash R)$. The partition induced by $M'$ has $S_{i+1}$
in the first class and is therefore different than $(R|S \backslash R)$,
contrary to the assumption of strong stability.

The second part of the proof is to show that every maxset of $(S^\wedge|\cup
S)$ is also a maxset of a strongly stable partition of $S$ after the formulae
$\wedge S_i$ are removed from it. This is achieved by showing that the
partition induced by the maxset is strongly stable.

Let $M$ be a maxset of $(S^\wedge|\cup S)$. Let $M' = M \cap S^\wedge$ and $M''
= M \cap \cup S$. Since $M$ is maximally consistent, $\wedge S_i \in M'$, $M''
\cup S_i \not\models \bot$ and $S_i \subseteq M''$ are equivalent to each
other. This proves that $M$ and $M''$ are equivalent.

The partition induced by $M''$ on $S$ is $(R|S \backslash R)$, where $R=\{S_i
~|~ M'' \cup S_i \not\models \bot\}$, which can be rewritten as $R=\{S_i ~|~
\wedge S_i \in M'\}$. Since $M'$ is consistent, so is $R$.

This partition is not strongly stable only if it has a maxset $M'''$ that
induces a different partition. By Lemma~\ref{consistent-highest}, since $R$ is
consistent $M'''$ contains all of it. Therefore, the partition induced by
$M'''$ has all of $R$ in the first class. This partition can be different than
$(R|S \backslash R)$ only if some $S_i \in S \backslash R$ is in its first
class, which requires it to be consistent with $M'''$, and therefore with $R$.
This implies that the formulae of $S^\wedge$ that are consistent with $M'''$
form a strict superset of those consistent with $M'$, contrary to the
assumption that $M$ is a maxset.~\qed

\section{Unreliable sources}
\label{unreliable}

The distinction between reliable and unreliable source may be too crude: among
the unreliable sources, some may provide useful information while others do
not. This observation leads to partitions of three classes: reliable sources,
unreliable sources and source providing only false information.

The same fixpoint definition of the previous sections can be used also in this
case: a three-class ordered partition is assumed, and its maxsets are found;
each induces a partition, where the first class is defined as before, but the
second comprises only the sources that provide some formulae that are
consistent with the maxset. The third class contains all other sources, whose
formulae are all inconsistent with the maxset.

Weak and strong stability are defined as before: a partition is weakly or
strongly stable if one or all of its maxsets induce the partition itself.

The only difference between the previous cases is the definition of induced
maxset, which is now composed of three classes. The rest of the framework is as
before.

\begin{definition}

A set of formulae $M$ induces on the set of sources $S = \{S_1,\ldots,S_n\}$
the tripartition $(R|P|U)$ where

\begin{eqnarray*}
R &=&
\{S_i ~|~ M \not\models \neg A \mbox{ for all } A \in S_i\}	
\\
P &=&
\{S_i ~|~ M \not\models \neg A \mbox{ for some but not all } A \in S_i\}
\\
U &=&
\{S_i ~|~ M \models \neg A \mbox{ for all } A \in S_i\}
\end{eqnarray*}

\end{definition}

This definition can be used in the same fixpoint condition of the previous
sections, so that a three-class partition is stable if its maxsets induce the
partition itself. Depending on whether all or some of these maxsets do,
stability is strong or weak. Theorem~\ref{reachable} still holds: considering
only partitions that are reachable from the one in which all sources are
compared the same is not a restriction. In a further section even more refined
classifications of the sources are considered.

The results in this section concern ordered partition of three sources. Some
relate partitions of two classes and partitions of three. To avoid confusion,
these are called {\em bipartitions} and {\em tripartitions}, respectively.
Given the number of theorems in this section, a summary is in order:

\begin{itemize}

\item the tripartition induced by a plain maxset has it as a maxset;

\item weakly stable bipartitions correspond to weakly stable tripartitions and
vice versa;

\item strongly stable tripartitions may not correspond to strongly stable
bipartitions;

\item some weakly stable bipartitions can be turned into strongly stable
tripartitions, but not all;

\item the maxsets of weakly stable tripartitions contain all of $R$, like for
bipartitions; this however does not extend to $P$ nor to $U$; it is also not
always the case that if $M$ is a maxset of $(R|P|U)$ not inducing the same
partition, then $M \cap R$ is strictly larger than $R$ (this was the case for
bipartitions);

\item cycles of weakly stable tripartitions are possible; there are even cases
where this holds for all maxsets;

\item not all sets of sources have strongly stable partitions;

\item the relation linking weakly stable partitions and maxsets is not
transitive nor symmetric; the latter holds even if the partitions have the same
$R$; furthermore, the relation can be made to include an arbitrary graph of
partitions.

\end{itemize}

The first result carries Lemma~\ref{itself} from bipartitions to tripartitions.

\begin{theorem}
\label{tri-itself}

If $M$ is a plain maxset of $S$,
it is also a maxset of its induced tripartition.

\end{theorem}

\proof Let $(R|P|U)$ be the partition induced by $M$. By definition, $M$ is a
maxset of this tripartition if it is a maxset of $R \cup P \cup U$, its
intersection with $R \cup P$ a maxset of $R \cup P$ and its intersection with
$R$ a maxset of $R$. The first holds by assumption, as $M$ is a plain maxset.
For the other two conditions, since $M$ is consistent only maximal consistency
within $R \cup P$ and $R$ is left to prove.

Regarding $R \cup P$, by definition of induced partition $U=\{S_i \mid \forall
kb \in S_i ~.~ M \models \neg kb\}$. Since $M$ is consistent, it does not
contain any formula in $U$ and therefore only comprises formulae of the other
two classes: $M \cap \cup \cup (R \cup P)=M$. If $M \cap \cup (R \cup P)
\subset M' \cap \cup (R \cup P)$ for some $M'$ then $M \subset M'$; therefore,
$M'$ is inconsistent because $M$ is a plain maxset of $S$.

Regarding $R$, by definition of induced partition $R=\{S_i \mid \forall kb \in
S_i ~.~ M \not\models \neg kb\}$. Since $M$ is a plain maxset, if it is
consistent with $kb$ then it includes it, otherwise it would not be a maximally
consistent subset. As a result, $M$ contains all formulae from the sources in
$R$. Therefore, it is a maxset of $\cup R$ simply because it contains the whole
of it.~\qed

Plain maxsets are also maxsets of a partition if they obey a simple condition.

\begin{lemma}
\label{all-none}

If a plain maxset of $R \cup P \cup U$ contains all formulae in $R$ and none in
$U$, it is a maxset of the partition $(R|P|U)$.

\end{lemma}

\proof The claim requires $M \cap R$ to be a maxset of $R$ and $M \cap (R \cup
P)$ to be a maxset of $R \cup P$. For the first condition: $M \cap R$ is
consistent because $M$ is consistent (being a plain maxset); it is maximal
because it contains all of $R$.

The second condition also holds. Since $M \subseteq R \cup P \cup U$ (because
$M$ is a plain maxset of $R \cup P \cup U$) and $M \cap U = \emptyset$ (by
assumption), it holds $M \subseteq R \cup P$. Therefore, $M \cap (R \cup P) =
M$. Since $M$ is maximally consistent in $R \cup P \cup U$, is also maximally
consistent in $R \cup P$.~\qed

The converse is not always the case. A maxset may contain formulae in the third
class, as for example $\{x,y,z\}$ is a maxset of $(\{x\}|\{y\}|\{z\})$. This
may even happen for weakly stable partitions, where at least a maxset has to
contain no formula in the third class by definition.

\begin{lemma}

Some weakly stable tripartitions have maxsets that contain formulae in their
third class.

\end{lemma}

\proof The maxsets of the the partition
{} $(
{}    \{x \vee y\} |
{}    \{\neg x \wedge \neg z, \neg y\} |
{}    \{z\}
{} )$
are
{} $\{x \vee y, \neg x \wedge \neg z\}$ and
{} $\{x \vee y, \neg y, z\}$.
The first induces the partition itself, proving that weakly stable. The second
contains the formula in the third class.~\qed

This result relies on the definition of maxset of a partition, and in
particular on the possibility for a maxset to include formulae in the last
class if they do not contradict the rest of the maxset. Such an event may be an
hint that the partition is not sensible, as it classified as unreliable some
source providing acceptable information. This motivates the definition of
strongly stable partitions, where this cannot happen. Also, a different
definition of unreliability is considered in a following section.

As far as weak stability is concerned, the introduction of the third class does
not change anything: weakly stable bipartitions and weakly stable tripartitions
are the same.

\begin{theorem}

For each weakly stable bipartition $(R|S \backslash R)$ there exists a weakly
stable tripartition $(R|P|U)$ where $P \cup U=S \backslash R$, and vice versa.

\end{theorem}

\proof If $(R|S \backslash R)$ is weakly stable, it has a maxset $M$ that
induces the bipartition itself. The tripartition induced by $M$ has the same
first class of the bipartition because the definition of the first class is the
same for bipartitions and tripartitions. Let $(R|P|U)$ be this tripartition.
Since $M$ is also a plain maxset, $(R|P|U)$ has $M$ as a maxset by
Lemma~\ref{tri-itself}. Since $M$ induces $(R|P|U)$, this tripartition is
weakly stable.

The other direction is proved in the same way. Let $(R|P|U)$ be a weakly stable
tripartition, and $M$ be a maxset that induces the tripartition itself. The
tripartition induced by $M$ is $(R|P|U)$; since the first class of induced
bipartitions and tripartitions are the same, the bipartition induced by $M$ is
$(R|S \backslash R)$. By Lemma~\ref{itself}, this bipartition has $M$ as a
maxset, which proves it to be weakly stable.~\qed

While weakly stable bipartitions and tripartition are the same, apart from the
split of the second class, the strongly stable bipartitions and tripartitions
exhibit quite a difference. Indeed, a bipartition may not be strongly stable
because of a maxset that induces a different bipartition; the addition of the
class of unreliable sources may block such a maxset.

\begin{theorem}

There exists a strongly stable tripartition $(R|P|U)$ such that $(R|P \cup U)$
is weakly but not strongly stable.

\end{theorem}

\proof The claim is proved on the same sources of Example~\ref{not-strong}:
{} $S_1=\{A\}$,
{} $S_2=\{B,C\}$, and
{} $S_3=\{D\}$,
with maxsets $\{A,B\}$, $\{B,C\}$ and $\{A,D\}$.

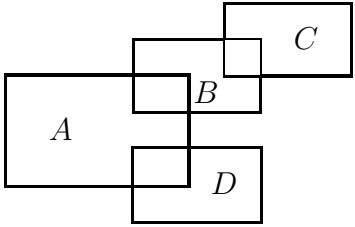
\setlength{\unitlength}{5000sp}%
\begingroup\makeatletter\ifx\SetFigFont\undefined%
\gdef\SetFigFont#1#2#3#4#5{%
  \reset@font\fontsize{#1}{#2pt}%
  \fontfamily{#3}\fontseries{#4}\fontshape{#5}%
  \selectfont}%
\fi\endgroup%
\begin{picture}(1754,1124)(4389,-3503)
\thinlines
{\color[rgb]{0,0,0}\put(5041,-2941){\framebox(630,360){}}
}%
{\color[rgb]{0,0,0}\put(5491,-2761){\framebox(630,360){}}
}%
\thicklines
{\color[rgb]{0,0,0}\put(4411,-3301){\framebox(900,540){}}
}%
{\color[rgb]{0,0,0}\put(5041,-3481){\framebox(630,360){}}
}%
{\color[rgb]{0,0,0}\put(5491,-2401){\line( 0,-1){180}}
\put(5491,-2581){\line(-1, 0){450}}
\put(5041,-2581){\line( 0,-1){360}}
\put(5041,-2941){\line( 1, 0){630}}
\put(5671,-2941){\line( 0, 1){180}}
\put(5671,-2761){\line( 1, 0){450}}
\put(6121,-2761){\line( 0, 1){360}}
\put(6121,-2401){\line(-1, 0){630}}
}%
\put(4681,-3076){\makebox(0,0)[b]{\smash{{\SetFigFont{12}{24.0}{\rmdefault}{\mddefault}{\itdefault}{\color[rgb]{0,0,0}$A$}%
}}}}
\put(5401,-2896){\makebox(0,0)[b]{\smash{{\SetFigFont{12}{24.0}{\rmdefault}{\mddefault}{\itdefault}{\color[rgb]{0,0,0}$B$}%
}}}}
\put(5896,-2626){\makebox(0,0)[b]{\smash{{\SetFigFont{12}{24.0}{\rmdefault}{\mddefault}{\itdefault}{\color[rgb]{0,0,0}$C$}%
}}}}
\put(5491,-3346){\makebox(0,0)[b]{\smash{{\SetFigFont{12}{24.0}{\rmdefault}{\mddefault}{\itdefault}{\color[rgb]{0,0,0}$D$}%
}}}}
\end{picture}%
\nop{
             +--------+
             |     C  |
       +-----+...     |
       | B   .  .     |
       |     ...+-----+
+------+---+    |
|      +---+----+
|  A       |
|      +---+----+
+------|---+    |
       |        |
       |     D  |
       +--------+
}

As already proved, the bipartition $(A|(BC)D)$ is weakly but not strongly
stable because of the maxset $\{A,D\}$, which induces the different partition
$(AD|(BC))$. As a result, its other maxset $\{A,B\}$ is not a maxset of any
strongly stable partition, since the partition it induces is $(A|(BC)D)$, which
is not strongly stable.

The bipartition $(A|(BC)D)$ is related to the tripartition $(A|(BC)|D)$ as
specified by the claim: the set of all sources and the first class are the
same. The tripartition $(A|(BC)|D)$ is strongly stable, as its only maxset
$\{A,B\}$ induces the tripartition itself: it is consistent with $A$, it is
consistent with one but not all formulae in $\{B,C\}$, and is inconsistent with
$D$. This proves that a bipartition that is not strongly stable can be turned
into a strongly stable tripartition.~\qed

While some bipartitions can be turned from weakly to strongly stable by
splitting their second class, this is not the case for all of them.

\begin{theorem}

There exists a weakly stable tripartition $(R|P|U)$ such that $(R|P \cup U)$ is
weakly but not strongly stable.

\end{theorem}

\proof The sources are $S=\{A, (BC), (DE)\}$ and the plain maxsets of $S$ are
$\{A,B,E\}$, $\{B,C\}$ and $\{A,D,E\}$.

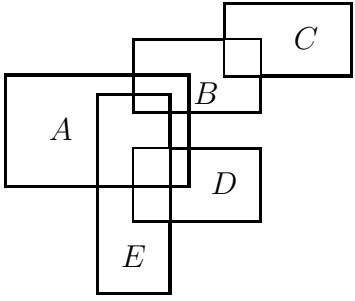
\setlength{\unitlength}{5000sp}%
\begingroup\makeatletter\ifx\SetFigFont\undefined%
\gdef\SetFigFont#1#2#3#4#5{%
  \reset@font\fontsize{#1}{#2pt}%
  \fontfamily{#3}\fontseries{#4}\fontshape{#5}%
  \selectfont}%
\fi\endgroup%
\begin{picture}(1754,1484)(4389,-3863)
\thinlines
{\color[rgb]{0,0,0}\put(5041,-2941){\framebox(630,360){}}
}%
{\color[rgb]{0,0,0}\put(5041,-3481){\framebox(630,360){}}
}%
{\color[rgb]{0,0,0}\put(5491,-2761){\framebox(630,360){}}
}%
{\color[rgb]{0,0,0}\put(4861,-3841){\framebox(360,990){}}
}%
\thicklines
{\color[rgb]{0,0,0}\put(4411,-3301){\framebox(900,540){}}
}%
{\color[rgb]{0,0,0}\put(5491,-2581){\line(-1, 0){450}}
\put(5041,-2581){\line( 0,-1){360}}
\put(5041,-2941){\line( 1, 0){630}}
\put(5671,-2941){\line( 0, 1){180}}
\put(5671,-2761){\line( 1, 0){450}}
\put(6121,-2761){\line( 0, 1){360}}
\put(6121,-2401){\line(-1, 0){630}}
\put(5491,-2401){\line( 0,-1){180}}
}%
{\color[rgb]{0,0,0}\put(5221,-2851){\line( 0,-1){270}}
\put(5221,-3121){\line( 1, 0){450}}
\put(5671,-3121){\line( 0,-1){360}}
\put(5671,-3481){\line(-1, 0){450}}
\put(5221,-3481){\line( 0,-1){360}}
\put(5221,-3841){\line(-1, 0){360}}
\put(4861,-3841){\line( 0, 1){990}}
\put(4861,-2851){\line( 1, 0){360}}
}%
\put(4681,-3076){\makebox(0,0)[b]{\smash{{\SetFigFont{12}{24.0}{\rmdefault}{\mddefault}{\itdefault}{\color[rgb]{0,0,0}$A$}%
}}}}
\put(5401,-2896){\makebox(0,0)[b]{\smash{{\SetFigFont{12}{24.0}{\rmdefault}{\mddefault}{\itdefault}{\color[rgb]{0,0,0}$B$}%
}}}}
\put(5896,-2626){\makebox(0,0)[b]{\smash{{\SetFigFont{12}{24.0}{\rmdefault}{\mddefault}{\itdefault}{\color[rgb]{0,0,0}$C$}%
}}}}
\put(5491,-3346){\makebox(0,0)[b]{\smash{{\SetFigFont{12}{24.0}{\rmdefault}{\mddefault}{\itdefault}{\color[rgb]{0,0,0}$D$}%
}}}}
\put(5041,-3706){\makebox(0,0)[b]{\smash{{\SetFigFont{12}{24.0}{\rmdefault}{\mddefault}{\itdefault}{\color[rgb]{0,0,0}$E$}%
}}}}
\end{picture}%
\nop{
               +--------+
               |     C  |
       +-------+...     |
       |   B   .  .     |
       |       ...+-----+
+------+-----+    |  
|      |     |    |
|  +---+--+  |    |
|  |   +--+--+----+
|A |      |  |  
|  |   ...+--+----+
|  |   .  .  |    |
+--+---+--+--+    |  
   |   .  .    D  |
   |   ...+-------+
   | E    |
   +------+
}

The bipartition $(A|(BC)(DE))$ is weakly but not strongly stable. Indeed, it
has two maxsets: $\{A,B,E\}$ and $\{A,D,E\}$. The first induces the partition
itself, proving it weakly stable. The second induces $(A(DE)|(BC))$, proving
that stability is not strong.

The maxset $\{A,B,E\}$ induces the tripartition $(A|(BC)(DE)|)$. This
tripartition has two maxsets: $\{A,B,E\}$ and $\{A,D,E\}$. The second induces
the different tripartition $(A(DE)||(BC))$. As a result, the tripartition is
weakly but not strongly stable.~\qed

Finally, some strongly stable bipartition cannot be turned into a strongly
stable tripartition, no matter how the second class is split. It will indeed be
proved that some sets of sources do not have any strongly stable tripartition,
while strongly stable bipartitions are guaranteed to exist as a consequence of
Theorem~\ref{maximal} or of Theorem~\ref{partition-of-strong}. Such a set is
provided in the following Corollary~\ref{no-strong}.

Theorem~\ref{allp} proves a number of results about the first class of weakly
stable bipartitions: they have a consistent first class, their maxsets contain
all of it, and the partition they induce have a larger or equal first class.
All of this extend to tripartitions because the proof was based on the first
class, which is defined in the same way. However, nothing like this holds for
the second and the third class of a tripartition.

\begin{theorem}

There exists a tripartition $(R|P|U)$ such that one of its maxsets induces a
partition $(R'|P'|U')$ such that
{} $P \not\subseteq P'$,
{} $U \not\subseteq U'$ and
{} $R \cup P \not\subseteq R' \cup P'$.

\end{theorem}

\proof The claim is proved using the sources $\{A,B,C\}$, $\{D,E\}$ and $\{F\}$
with the plain maxsets $\{A,B,C\}$, $\{B,C,D\}$ and $\{D,E,F\}$.

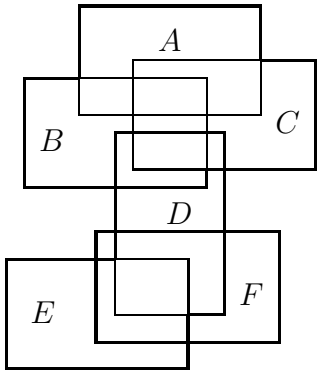
\setlength{\unitlength}{5000sp}%
\begingroup\makeatletter\ifx\SetFigFont\undefined%
\gdef\SetFigFont#1#2#3#4#5{%
  \reset@font\fontsize{#1}{#2pt}%
  \fontfamily{#3}\fontseries{#4}\fontshape{#5}%
  \selectfont}%
\fi\endgroup%
\begin{picture}(1574,1844)(4479,-4403)
\thinlines
{\color[rgb]{0,0,0}\put(4591,-3481){\framebox(900,540){}}
}%
{\color[rgb]{0,0,0}\put(4861,-3121){\framebox(900,540){}}
}%
{\color[rgb]{0,0,0}\put(5131,-3391){\framebox(900,540){}}
}%
{\color[rgb]{0,0,0}\put(5041,-4111){\framebox(540,900){}}
}%
{\color[rgb]{0,0,0}\put(4501,-4381){\framebox(900,540){}}
}%
\thicklines
{\color[rgb]{0,0,0}\put(5491,-3391){\line( 0,-1){ 90}}
\put(5491,-3481){\line(-1, 0){900}}
\put(4591,-3481){\line( 0, 1){540}}
\put(4591,-2941){\line( 1, 0){270}}
\put(4861,-2941){\line( 0, 1){360}}
\put(4861,-2581){\line( 1, 0){900}}
\put(5761,-2581){\line( 0,-1){270}}
\put(5761,-2851){\line( 1, 0){270}}
\put(6031,-2851){\line( 0,-1){540}}
\put(6031,-3391){\line(-1, 0){540}}
}%
{\color[rgb]{0,0,0}\put(5041,-3211){\line( 1, 0){540}}
\put(5581,-3211){\line( 0,-1){900}}
\put(5581,-4111){\line(-1, 0){180}}
\put(5401,-4111){\line( 0,-1){270}}
\put(5401,-4381){\line(-1, 0){900}}
\put(4501,-4381){\line( 0, 1){540}}
\put(4501,-3841){\line( 1, 0){540}}
\put(5041,-3841){\line( 0, 1){630}}
}%
{\color[rgb]{0,0,0}\put(4951,-4246){\framebox(900,540){}}
}%
\put(5311,-2806){\makebox(0,0)[b]{\smash{{\SetFigFont{12}{24.0}{\rmdefault}{\mddefault}{\itdefault}{\color[rgb]{0,0,0}$A$}%
}}}}
\put(4726,-3301){\makebox(0,0)[b]{\smash{{\SetFigFont{12}{24.0}{\rmdefault}{\mddefault}{\itdefault}{\color[rgb]{0,0,0}$B$}%
}}}}
\put(5896,-3211){\makebox(0,0)[b]{\smash{{\SetFigFont{12}{24.0}{\rmdefault}{\mddefault}{\itdefault}{\color[rgb]{0,0,0}$C$}%
}}}}
\put(5356,-3661){\makebox(0,0)[b]{\smash{{\SetFigFont{12}{24.0}{\rmdefault}{\mddefault}{\itdefault}{\color[rgb]{0,0,0}$D$}%
}}}}
\put(4681,-4156){\makebox(0,0)[b]{\smash{{\SetFigFont{12}{24.0}{\rmdefault}{\mddefault}{\itdefault}{\color[rgb]{0,0,0}$E$}%
}}}}
\put(5716,-4066){\makebox(0,0)[b]{\smash{{\SetFigFont{12}{24.0}{\rmdefault}{\mddefault}{\itdefault}{\color[rgb]{0,0,0}$F$}%
}}}}
\end{picture}%
\nop{
          +-----------------+
          |   A      .......+-----+
     +----+.............    .   C |
     |    +.................+     |
     | B       +---------+        |
     |         |     . . |        |
     |         |     ..+----------+
     +-----------------+ |
               | D       |
            +--+---------+-----+
            |  |         |  F  |
    +-------+--+......   |     |
    |  E    |  .     .   |     |
    |       |  .......---+     |
    |       +------------------+
    +----------------+
}

A weakly stable partition of these sources is $(|(ABC)(DE)|F)$, since its
maxset $\{B,C,D\}$ induces the partition itself.

This partition also has the two maxsets $\{A,B,C\}$ and $\{D,E,F\}$. The first
maxset $\{A,B,C\}$ induces $((ABC)||(DE)F)$, which shows that the second class
does not necessarily increase, neither does its union with the first. The
second maxset $\{D,E,F\}$ induces $((DE)F||(ABC))$, whose third class is
incomparable with that of the original partition.~\qed

Incidentally, the set of sources has a strongly stable tripartition:
$((ABC)||(DE)F)$. Indeed, the only maxset of this tripartition is $\{A,B,C\}$,
which induces the partition itself.

In this and in the previous proof, going from a partition to a maxset and to
its induced partition reduces the second class. This is however easy to show
not to always be the case, with the same sources but $\{F,G\}$ in place of
$\{F\}$, where $G$ is only consistent with $F$. Thanks to this change, the
maxset $\{D,E,F\}$ induces the partition $((DE)|(FG)|(ABC))$.

Another common feature of the above proofs is that the tripartitions that are
weakly but not strongly stable have a maxset whose induced partition has a
second class that is strictly larger than the original partition. This is
however not always the case.

\begin{theorem}

There exists a partition with two maxsets such that the first induces the
partition itself and the second induces a partition with a smaller second
class.

\end{theorem}

\proof The proof is based on the partition $(A|(BCD)(EF)|)$ whose plain
maxsets are $\{A,B,C\}$, $\{A,B,E\}$, $\{B,C,D\}$ and $\{E,F\}$.

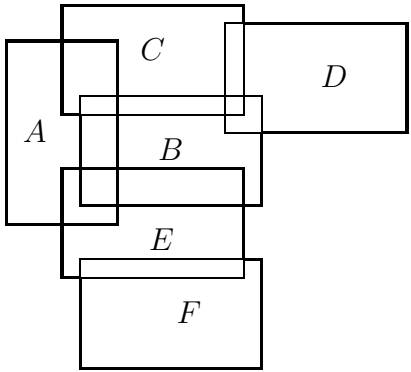
\setlength{\unitlength}{5000sp}%
\begingroup\makeatletter\ifx\SetFigFont\undefined%
\gdef\SetFigFont#1#2#3#4#5{%
  \reset@font\fontsize{#1}{#2pt}%
  \fontfamily{#3}\fontseries{#4}\fontshape{#5}%
  \selectfont}%
\fi\endgroup%
\begin{picture}(2024,1844)(4299,-4403)
\thinlines
{\color[rgb]{0,0,0}\put(4591,-3121){\framebox(900,540){}}
}%
{\color[rgb]{0,0,0}\put(5401,-3211){\framebox(900,540){}}
}%
{\color[rgb]{0,0,0}\put(4681,-3571){\framebox(900,540){}}
}%
{\color[rgb]{0,0,0}\put(4591,-3931){\framebox(900,540){}}
}%
{\color[rgb]{0,0,0}\put(4681,-4381){\framebox(900,540){}}
}%
\thicklines
{\color[rgb]{0,0,0}\put(4321,-3661){\framebox(540,900){}}
}%
{\color[rgb]{0,0,0}\put(4681,-3121){\line(-1, 0){ 90}}
\put(4591,-3121){\line( 0, 1){540}}
\put(4591,-2581){\line( 1, 0){900}}
\put(5491,-2581){\line( 0,-1){ 90}}
\put(5491,-2671){\line( 1, 0){810}}
\put(6301,-2671){\line( 0,-1){540}}
\put(6301,-3211){\line(-1, 0){720}}
\put(5581,-3211){\line( 0,-1){360}}
\put(5581,-3571){\line(-1, 0){900}}
\put(4681,-3571){\line( 0, 1){450}}
}%
{\color[rgb]{0,0,0}\put(4681,-3931){\line(-1, 0){ 90}}
\put(4591,-3931){\line( 0, 1){540}}
\put(4591,-3391){\line( 1, 0){900}}
\put(5491,-3391){\line( 0,-1){450}}
\put(5491,-3841){\line( 1, 0){ 90}}
\put(5581,-3841){\line( 0,-1){540}}
\put(5581,-4381){\line(-1, 0){900}}
\put(4681,-4381){\line( 0, 1){450}}
}%
\put(5041,-2851){\makebox(0,0)[b]{\smash{{\SetFigFont{12}{24.0}{\rmdefault}{\mddefault}{\itdefault}{\color[rgb]{0,0,0}$C$}%
}}}}
\put(5941,-2986){\makebox(0,0)[b]{\smash{{\SetFigFont{12}{24.0}{\rmdefault}{\mddefault}{\itdefault}{\color[rgb]{0,0,0}$D$}%
}}}}
\put(5131,-3346){\makebox(0,0)[b]{\smash{{\SetFigFont{12}{24.0}{\rmdefault}{\mddefault}{\itdefault}{\color[rgb]{0,0,0}$B$}%
}}}}
\put(5221,-4156){\makebox(0,0)[b]{\smash{{\SetFigFont{12}{24.0}{\rmdefault}{\mddefault}{\itdefault}{\color[rgb]{0,0,0}$F$}%
}}}}
\put(4456,-3256){\makebox(0,0)[b]{\smash{{\SetFigFont{12}{24.0}{\rmdefault}{\mddefault}{\itdefault}{\color[rgb]{0,0,0}$A$}%
}}}}
\put(5086,-3796){\makebox(0,0)[b]{\smash{{\SetFigFont{12}{24.0}{\rmdefault}{\mddefault}{\itdefault}{\color[rgb]{0,0,0}$E$}%
}}}}
\end{picture}%
\nop{
        /--- C ---\                  .
       |           +--- D
 A ----+---- B ---/
  \        /
    \    /
      \/
       +----- E ----+---- F
}

One of the maxsets of the tripartition $(A|(BCD)(EF)|)$ is $\{A,B,E\}$, which
induces the partition itself. This is the first part of the claim. The only
other maxset of the tripartition is $\{A,B,C\}$, which induces
$(A|(BCD)|(EF))$. This partition has the same first class of the original one,
but a smaller second class.~\qed

\begin{theorem}

There exists a partition with two maxsets such that the first induces the
partition itself and the second induces a partition with a smaller second
class.

\end{theorem}

For bipartitions, a strongly stable partition can always be found by starting
from a maxset and then following the induced partition and then its maxsets, or
starting from a weakly stable partition and following a similar path. Nothing
like this can be done on tripartitions: one may start from a weakly stable one,
follow the maxsets and then the induced partitions and so only only to end up
in the original tripartition. This can be the case even for all maxsets of a
tripartition, as the following example shows.

\begin{lemma}
\label{two-two}

There exists two tripartitions $T$ and $T'$ that have the same first class and
the same two maxsets $M$ and $M'$, where $M$ induces $T$ and $M'$ induces $T'$.

\end{lemma}

\proof The sources are $\{A\}$, $\{B,C\}$, $\{D,E\}$, $\{F,G\}$ and their plain
maxsets are $\{B,C\}$, $\{A,B,D\}$, $\{D,E\}$, $\{A,E,F\}$ and $\{F,G\}$.

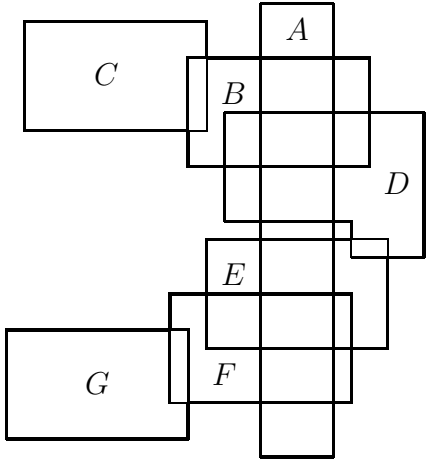
\setlength{\unitlength}{5000sp}%
\begingroup\makeatletter\ifx\SetFigFont\undefined%
\gdef\SetFigFont#1#2#3#4#5{%
  \reset@font\fontsize{#1}{#2pt}%
  \fontfamily{#3}\fontseries{#4}\fontshape{#5}%
  \selectfont}%
\fi\endgroup%
\begin{picture}(2114,2294)(4299,-4763)
\thinlines
{\color[rgb]{0,0,0}\put(4411,-3121){\framebox(900,540){}}
}%
{\color[rgb]{0,0,0}\put(5221,-3301){\framebox(900,540){}}
}%
{\color[rgb]{0,0,0}\put(5401,-3031){\line( 0,-1){540}}
\put(5401,-3571){\line( 1, 0){630}}
\put(6031,-3571){\line( 0,-1){180}}
\put(6031,-3751){\line( 1, 0){360}}
\put(6391,-3751){\line( 0, 1){720}}
\put(6391,-3031){\line(-1, 0){990}}
}%
{\color[rgb]{0,0,0}\put(5131,-4471){\framebox(900,540){}}
}%
{\color[rgb]{0,0,0}\put(4321,-4651){\framebox(900,540){}}
}%
\thicklines
{\color[rgb]{0,0,0}\put(5581,-2491){\line( 0,-1){2250}}
\put(5581,-4741){\line( 1, 0){360}}
\put(5941,-4741){\line( 0, 1){2250}}
\put(5941,-2491){\line(-1, 0){360}}
}%
{\color[rgb]{0,0,0}\put(5131,-4111){\line(-1, 0){810}}
\put(4321,-4111){\line( 0,-1){540}}
\put(4321,-4651){\line( 1, 0){900}}
\put(5221,-4651){\line( 0, 1){180}}
\put(5221,-4471){\line( 1, 0){810}}
\put(6031,-4471){\line( 0, 1){540}}
\put(6031,-3931){\line(-1, 0){900}}
\put(5131,-3931){\line( 0,-1){180}}
}%
\thinlines
{\color[rgb]{0,0,0}\put(5311,-4201){\framebox(900,540){}}
}%
\thicklines
{\color[rgb]{0,0,0}\put(5311,-3661){\line( 1, 0){720}}
\put(6031,-3661){\line( 0, 1){ 90}}
\put(6031,-3571){\line(-1, 0){630}}
\put(5401,-3571){\line( 0, 1){540}}
\put(5401,-3031){\line( 1, 0){990}}
\put(6391,-3031){\line( 0,-1){720}}
\put(6391,-3751){\line(-1, 0){180}}
\put(6211,-3751){\line( 0,-1){450}}
\put(6211,-4201){\line(-1, 0){900}}
\put(5311,-4201){\line( 0, 1){540}}
}%
{\color[rgb]{0,0,0}\put(5221,-3121){\line(-1, 0){810}}
\put(4411,-3121){\line( 0, 1){540}}
\put(4411,-2581){\line( 1, 0){900}}
\put(5311,-2581){\line( 0,-1){180}}
\put(5311,-2761){\line( 1, 0){810}}
\put(6121,-2761){\line( 0,-1){540}}
\put(6121,-3301){\line(-1, 0){900}}
\put(5221,-3301){\line( 0, 1){180}}
}%
\put(5761,-2671){\makebox(0,0)[b]{\smash{{\SetFigFont{12}{24.0}{\rmdefault}{\mddefault}{\itdefault}{\color[rgb]{0,0,0}$A$}%
}}}}
\put(5446,-2986){\makebox(0,0)[b]{\smash{{\SetFigFont{12}{24.0}{\rmdefault}{\mddefault}{\itdefault}{\color[rgb]{0,0,0}$B$}%
}}}}
\put(4816,-2896){\makebox(0,0)[b]{\smash{{\SetFigFont{12}{24.0}{\rmdefault}{\mddefault}{\itdefault}{\color[rgb]{0,0,0}$C$}%
}}}}
\put(5401,-4381){\makebox(0,0)[b]{\smash{{\SetFigFont{12}{24.0}{\rmdefault}{\mddefault}{\itdefault}{\color[rgb]{0,0,0}$F$}%
}}}}
\put(4771,-4426){\makebox(0,0)[b]{\smash{{\SetFigFont{12}{24.0}{\rmdefault}{\mddefault}{\itdefault}{\color[rgb]{0,0,0}$G$}%
}}}}
\put(5446,-3886){\makebox(0,0)[b]{\smash{{\SetFigFont{12}{24.0}{\rmdefault}{\mddefault}{\itdefault}{\color[rgb]{0,0,0}$E$}%
}}}}
\put(6256,-3436){\makebox(0,0)[b]{\smash{{\SetFigFont{12}{24.0}{\rmdefault}{\mddefault}{\itdefault}{\color[rgb]{0,0,0}$D$}%
}}}}
\end{picture}%
\nop{
C - B = A = D
        A   |
G - F = A = E
}

The tripartitions and the maxsets of the claim are:

\begin{eqnarray*}
T 	&=&	(A|(BC)(DE)|(FG))	\\
T'	&=&	(A|(FG)(DE)|(BC))	\\
M	&=&	\{A,B,D\}		\\
M'	&=&	\{A,E,F\}
\end{eqnarray*}

Both tripartitions have $A$ only in their first class. All their maxsets
therefore contain $A$. The only plain maxsets containing $A$ are $M$ and $M'$.
Both are maxsets of the tripartitions, since they cannot be consistently added
a formula of the second class even removing the formulae of the third ($F$ for
the first tripartition and $B$ for the second). That $M$ induces $T$ and $M'$
induces $T'$ is a consequence of the definition.~\qed

A first consequence of this lemma is the existence of strongly stable
bipartitions that cannot be turned into strongly stable tripartitions by
splitting their second class.

\begin{corollary}
\label{no-strong}

There exists a strongly stable bipartition $(R|S \backslash R)$ such that
$(R|P|U)$ is not strongly stable
for any $P$ and $U$ such that $R \cup P \cup U = S$.

\end{corollary}

\proof The tripartitions $T$ and $T$ of the previous theorem are weakly but not
strongly stable tripartition. Since their first classes coincide, they are
associated the same bipartition, which is strongly stable because its maxsets
$M$ and $M'$ induce the tripartition, and therefore induce the same
bipartition.~\qed

Lemma~\ref{two-two} also proves that a strongly stable tripartition cannot be
found by moving from tripartitions to maxsets and then to their induced
tripartitions. A strongly stable tripartition requires such a path being a loop
of length two (tripartition -- maxset -- same tripartition), while the lemma
shows a loop of length four.

Incidentally, the sources of the proof have a strongly stable tripartition:
$((BC)||A(DE)(FG))$; its only maxset is $\{B,C\}$, which induces the partition
itself. This means that the tripartition -- maxset -- tripartition procedure
may fail to find a strongly stable tripartition even if one exists. However, it
is also the case that a set of sources has no strongly stable tripartition.

\begin{theorem}

Some sets of sources have no strongly stable tripartitions.

\end{theorem}

\proof The claim is proved using four sources: $\{A,A'\}$, $\{B,B'\}$,
$\{C,C'\}$, and $\{D,D'\}$. The maxsets containing both $A$ and $A'$ are
$\{A,A',B,C\}$ and $\{A,A',C',D\}$; a similar pair of maxsets contain both $B$
and $B'$, another $C$ and $C'$ and yet another $D$ and $D'$.

Graphically, these formulae can be seen as a composition of four modules, one
for each source. The module for the first source is as follows, with the
smaller figure on the right being a simplified representation.

\

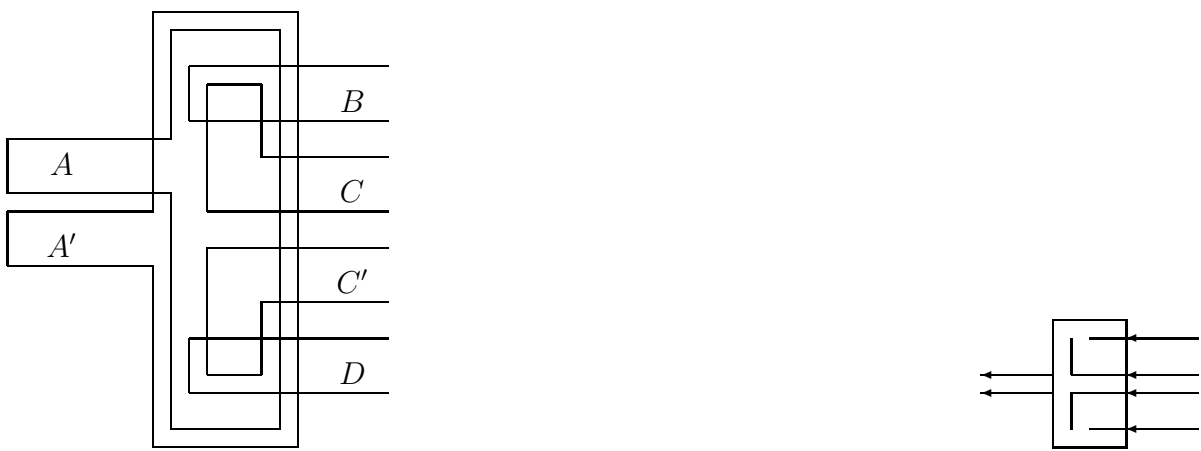
\ttytex{
\setlength{\unitlength}{5000sp}%
\begingroup\makeatletter\ifx\SetFigFont\undefined%
\gdef\SetFigFont#1#2#3#4#5{%
  \reset@font\fontsize{#1}{#2pt}%
  \fontfamily{#3}\fontseries{#4}\fontshape{#5}%
  \selectfont}%
\fi\endgroup%
\begin{picture}(1914,2184)(4309,-4843)
\thinlines
{\color[rgb]{0,0,0}\put(6211,-3391){\line(-1, 0){630}}
\put(5581,-3391){\line( 0, 1){360}}
\put(5581,-3031){\line(-1, 0){270}}
\put(5311,-3031){\line( 0,-1){630}}
\put(5311,-3661){\line( 1, 0){900}}
}%
{\color[rgb]{0,0,0}\put(6211,-2941){\line(-1, 0){990}}
\put(5221,-2941){\line( 0,-1){270}}
\put(5221,-3211){\line( 1, 0){990}}
}%
{\color[rgb]{0,0,0}\put(6211,-4111){\line(-1, 0){630}}
\put(5581,-4111){\line( 0,-1){360}}
\put(5581,-4471){\line(-1, 0){270}}
\put(5311,-4471){\line( 0, 1){630}}
\put(5311,-3841){\line( 1, 0){900}}
}%
{\color[rgb]{0,0,0}\put(6211,-4291){\line(-1, 0){990}}
\put(5221,-4291){\line( 0,-1){270}}
\put(5221,-4561){\line( 1, 0){990}}
}%
{\color[rgb]{0,0,0}\put(5671,-2761){\line( 0,-1){1980}}
\put(5671,-4741){\line(-1, 0){540}}
\put(5131,-4741){\line( 0, 1){1170}}
\put(5131,-3571){\line(-1, 0){810}}
\put(4321,-3571){\line( 0, 1){270}}
\put(4321,-3301){\line( 1, 0){810}}
\put(5131,-3301){\line( 0, 1){540}}
\put(5131,-2761){\line( 1, 0){540}}
}%
{\color[rgb]{0,0,0}\put(5041,-3661){\line(-1, 0){720}}
\put(4321,-3661){\line( 0,-1){270}}
\put(4321,-3931){\line( 1, 0){720}}
\put(5041,-3931){\line( 0,-1){900}}
\put(5041,-4831){\line( 1, 0){720}}
\put(5761,-4831){\line( 0, 1){2160}}
\put(5761,-2671){\line(-1, 0){720}}
\put(5041,-2671){\line( 0,-1){990}}
}%
\put(4591,-3481){\makebox(0,0)[b]{\smash{{\SetFigFont{12}{24.0}{\rmdefault}{\mddefault}{\itdefault}{\color[rgb]{0,0,0}$A$}%
}}}}
\put(4591,-3886){\makebox(0,0)[b]{\smash{{\SetFigFont{12}{24.0}{\rmdefault}{\mddefault}{\itdefault}{\color[rgb]{0,0,0}$A'$}%
}}}}
\put(6031,-3166){\makebox(0,0)[b]{\smash{{\SetFigFont{12}{24.0}{\rmdefault}{\mddefault}{\itdefault}{\color[rgb]{0,0,0}$B$}%
}}}}
\put(6031,-4516){\makebox(0,0)[b]{\smash{{\SetFigFont{12}{24.0}{\rmdefault}{\mddefault}{\itdefault}{\color[rgb]{0,0,0}$D$}%
}}}}
\put(6031,-3616){\makebox(0,0)[b]{\smash{{\SetFigFont{12}{24.0}{\rmdefault}{\mddefault}{\itdefault}{\color[rgb]{0,0,0}$C$}%
}}}}
\put(6031,-4066){\makebox(0,0)[b]{\smash{{\SetFigFont{12}{24.0}{\rmdefault}{\mddefault}{\itdefault}{\color[rgb]{0,0,0}$C'$}%
}}}}
\end{picture}%
\hfill
\setlength{\unitlength}{5000sp}%
\begingroup\makeatletter\ifx\SetFigFont\undefined%
\gdef\SetFigFont#1#2#3#4#5{%
  \reset@font\fontsize{#1}{#2pt}%
  \fontfamily{#3}\fontseries{#4}\fontshape{#5}%
  \selectfont}%
\fi\endgroup%
\begin{picture}(1104,654)(4849,-3763)
\thinlines
{\color[rgb]{0,0,0}\put(5941,-3211){\vector(-1, 0){360}}
}%
{\color[rgb]{0,0,0}\put(5941,-3391){\vector(-1, 0){360}}
}%
{\color[rgb]{0,0,0}\put(5941,-3481){\vector(-1, 0){360}}
}%
{\color[rgb]{0,0,0}\put(5941,-3661){\vector(-1, 0){360}}
}%
{\color[rgb]{0,0,0}\put(5221,-3751){\framebox(360,630){}}
}%
{\color[rgb]{0,0,0}\put(5221,-3391){\vector(-1, 0){360}}
}%
{\color[rgb]{0,0,0}\put(5221,-3481){\vector(-1, 0){360}}
}%
{\color[rgb]{0,0,0}\put(5581,-3391){\line(-1, 0){270}}
\put(5311,-3391){\line( 0, 1){180}}
}%
{\color[rgb]{0,0,0}\put(5581,-3481){\line(-1, 0){270}}
\put(5311,-3481){\line( 0,-1){180}}
}%
{\color[rgb]{0,0,0}\put(5401,-3211){\line( 1, 0){180}}
}%
{\color[rgb]{0,0,0}\put(5581,-3661){\line(-1, 0){180}}
}%
\end{picture}%
}{
[figure]
}

\

The modules are composed so that formulae do not intersect outside them. There
are two maxsets for each source. For the first, they are $\{A,A',B,C\}$ and
$\{A,A',C',D\}$. They induce the partitions $((AA')|(BB')(CC')|(DD'))$ and
$((AA'|(CC')(DD')|)$, respectively; both of them have both maxsets. As a
result, they are not strongly stable.

The claim requires that a set of sources can be build in a symmetric way, so
that what holds for a source holds for the others as well. The four modules,
one for each source, are deployed in circle:

\

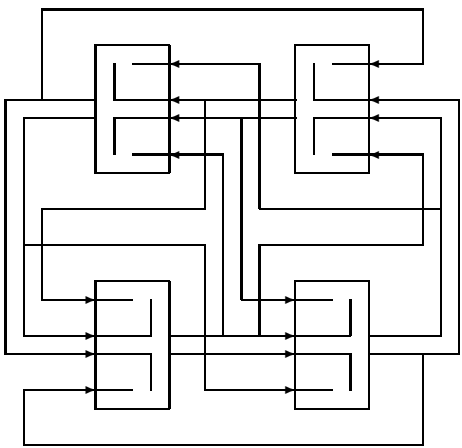
\setlength{\unitlength}{5000sp}%
\begingroup\makeatletter\ifx\SetFigFont\undefined%
\gdef\SetFigFont#1#2#3#4#5{%
  \reset@font\fontsize{#1}{#2pt}%
  \fontfamily{#3}\fontseries{#4}\fontshape{#5}%
  \selectfont}%
\fi\endgroup%
\begin{picture}(2274,2184)(3769,-5113)
\thinlines
{\color[rgb]{0,0,0}\put(5221,-3751){\framebox(360,630){}}
}%
{\color[rgb]{0,0,0}\put(5581,-3391){\line(-1, 0){270}}
\put(5311,-3391){\line( 0, 1){180}}
}%
{\color[rgb]{0,0,0}\put(5581,-3481){\line(-1, 0){270}}
\put(5311,-3481){\line( 0,-1){180}}
}%
{\color[rgb]{0,0,0}\put(5401,-3211){\line( 1, 0){180}}
}%
{\color[rgb]{0,0,0}\put(5581,-3661){\line(-1, 0){180}}
}%
{\color[rgb]{0,0,0}\put(4231,-3751){\framebox(360,630){}}
}%
{\color[rgb]{0,0,0}\put(4591,-3391){\line(-1, 0){270}}
\put(4321,-3391){\line( 0, 1){180}}
}%
{\color[rgb]{0,0,0}\put(4591,-3481){\line(-1, 0){270}}
\put(4321,-3481){\line( 0,-1){180}}
}%
{\color[rgb]{0,0,0}\put(4411,-3211){\line( 1, 0){180}}
}%
{\color[rgb]{0,0,0}\put(4591,-3661){\line(-1, 0){180}}
}%
{\color[rgb]{0,0,0}\put(5221,-4921){\framebox(360,630){}}
}%
{\color[rgb]{0,0,0}\put(5221,-4561){\line( 1, 0){270}}
\put(5491,-4561){\line( 0, 1){180}}
}%
{\color[rgb]{0,0,0}\put(5221,-4651){\line( 1, 0){270}}
\put(5491,-4651){\line( 0,-1){180}}
}%
{\color[rgb]{0,0,0}\put(5401,-4381){\line(-1, 0){180}}
}%
{\color[rgb]{0,0,0}\put(5221,-4831){\line( 1, 0){180}}
}%
{\color[rgb]{0,0,0}\put(4231,-4921){\framebox(360,630){}}
}%
{\color[rgb]{0,0,0}\put(4231,-4561){\line( 1, 0){270}}
\put(4501,-4561){\line( 0, 1){180}}
}%
{\color[rgb]{0,0,0}\put(4231,-4651){\line( 1, 0){270}}
\put(4501,-4651){\line( 0,-1){180}}
}%
{\color[rgb]{0,0,0}\put(4411,-4381){\line(-1, 0){180}}
}%
{\color[rgb]{0,0,0}\put(4231,-4831){\line( 1, 0){180}}
}%
{\color[rgb]{0,0,0}\put(5221,-3391){\vector(-1, 0){630}}
}%
{\color[rgb]{0,0,0}\put(5221,-3481){\vector(-1, 0){630}}
}%
{\color[rgb]{0,0,0}\put(4231,-3481){\line(-1, 0){360}}
\put(3871,-3481){\line( 0,-1){1080}}
\put(3871,-4561){\vector( 1, 0){360}}
}%
{\color[rgb]{0,0,0}\put(4231,-3391){\line(-1, 0){450}}
\put(3781,-3391){\line( 0,-1){1260}}
\put(3781,-4651){\vector( 1, 0){450}}
}%
{\color[rgb]{0,0,0}\put(4951,-3481){\line( 0,-1){900}}
\put(4951,-4381){\vector( 1, 0){270}}
}%
{\color[rgb]{0,0,0}\put(4591,-4561){\vector( 1, 0){630}}
}%
{\color[rgb]{0,0,0}\put(4591,-4651){\vector( 1, 0){630}}
}%
{\color[rgb]{0,0,0}\put(5581,-4561){\line( 1, 0){360}}
\put(5941,-4561){\line( 0, 1){1080}}
\put(5941,-3481){\vector(-1, 0){360}}
}%
{\color[rgb]{0,0,0}\put(5581,-4651){\line( 1, 0){450}}
\put(6031,-4651){\line( 0, 1){1260}}
\put(6031,-3391){\vector(-1, 0){450}}
}%
{\color[rgb]{0,0,0}\put(4861,-4561){\line( 0, 1){900}}
\put(4861,-3661){\vector(-1, 0){270}}
}%
{\color[rgb]{0,0,0}\put(3961,-3391){\line( 0, 1){450}}
\put(3961,-2941){\line( 1, 0){1890}}
\put(5851,-2941){\line( 0,-1){270}}
\put(5851,-3211){\vector(-1, 0){270}}
}%
{\color[rgb]{0,0,0}\put(5851,-4651){\line( 0,-1){450}}
\put(5851,-5101){\line(-1, 0){1980}}
\put(3871,-5101){\line( 0, 1){270}}
\put(3871,-4831){\vector( 1, 0){360}}
}%
{\color[rgb]{0,0,0}\put(4771,-3391){\line( 0,-1){540}}
\put(4771,-3931){\line(-1, 0){810}}
\put(3961,-3931){\line( 0,-1){450}}
\put(3961,-4381){\vector( 1, 0){270}}
}%
{\color[rgb]{0,0,0}\put(5041,-4561){\line( 0, 1){450}}
\put(5041,-4111){\line( 1, 0){810}}
\put(5851,-4111){\line( 0, 1){450}}
\put(5851,-3661){\vector(-1, 0){270}}
}%
{\color[rgb]{0,0,0}\put(5941,-3931){\line(-1, 0){900}}
\put(5041,-3931){\line( 0, 1){720}}
\put(5041,-3211){\vector(-1, 0){450}}
}%
{\color[rgb]{0,0,0}\put(3871,-4111){\line( 1, 0){900}}
\put(4771,-4111){\line( 0,-1){720}}
\put(4771,-4831){\vector( 1, 0){450}}
}%
\end{picture}%
\nop{
[figure]
}

\

The first module has $A$ and $A'$ as ``output'' and $B,C,C',D$ as ``input'';
the second module has $B$ and $B'$ as output and $C,D,D',A$ as input, etc. By
symmetry, the argument that holds for the maxsets $\{A,A',B,C\}$ and
$\{A,A',C',D\}$ carries to all other ones, showing that none of them induces a
tripartition that is strongly stable.~\qed

Given that strongly stable tripartitions do not always exist, a milder
condition can be defined. The proof of the theorem is based on four
tripartitions such that each induces not only itself but also one of the
others, forming a sort of cycle. While none induces only itself, one may
consider the group itself to be strongly stable, in the sense that the removal
of any element from it falsifies this condition.

Given two partitions $X$ and $Y$, that $X$ has a maxset that induces $Y$ is
denoted by $X \Rightarrow Y$. This relation among partitions allows rewriting
the definition of weakly and stably strong partitions as follows: $X
\Rightarrow X$ for weakly stable and $X \Rightarrow Y$ implies $Y = X$ for
strongly stable. The difference between bipartitions and tripartitions only
affects the definition of induced partition, which is part of the definition of
$\Rightarrow$.

\begin{definition}

The relation $\Rightarrow$ is defined as: given two partitions $X$ and $Y$, it
holds $X \Rightarrow Y$ if $X$ has a maxset that induces $Y$.

\end{definition}

The relation $\Rightarrow$ is defined over all partitions, including the ones
that are not weakly stable. For these, $X \Rightarrow X$ may not hold, meaning
that $\Rightarrow$ is not reflexive. Restricting to the weakly stable
partitions makes $\Rightarrow$ reflexive, but neither symmetric nor transitive,
as the following theorem proves. The relation is however serial: for every
tripartition $X$ it holds $X \Rightarrow Y$ for some possibly different
tripartition $Y$.

\begin{theorem}

The relation $\Rightarrow$ is serial but is not reflexive, not symmetric and
not transitive. When restricted to weakly stable tripartitions is reflexive but
neither symmetric nor transitive.

\end{theorem}

\proof The relation $\Rightarrow$ is serial if $X \Rightarrow Y$ holds for
every $X$ and some $Y$. This is the case because every tripartition $X$ has at
least a maxset $M$, and every maxset like $M$ induces a tripartition $Y$;
therefore, $X \Rightarrow Y$.

Not being reflexive is shown by the partition $(||A)$: its only maxset is
$\{A\}$, which induces $(A||)$. Restricting to weakly stable tripartitions
makes it reflexive by definition, since weak stability is defined as $X
\Rightarrow X$.

The relation $\Rightarrow$ is shown not symmetric and not transitive by the
following example.

\

\setlength{\unitlength}{5000sp}%
\begingroup\makeatletter\ifx\SetFigFont\undefined%
\gdef\SetFigFont#1#2#3#4#5{%
  \reset@font\fontsize{#1}{#2pt}%
  \fontfamily{#3}\fontseries{#4}\fontshape{#5}%
  \selectfont}%
\fi\endgroup%
\begin{picture}(2834,2184)(4659,-5023)
\thinlines
{\color[rgb]{0,0,0}\put(6571,-3661){\framebox(900,540){}}
}%
{\color[rgb]{0,0,0}\put(5851,-3481){\framebox(900,540){}}
}%
{\color[rgb]{0,0,0}\put(6031,-4921){\framebox(900,540){}}
}%
{\color[rgb]{0,0,0}\put(5401,-4741){\framebox(900,540){}}
}%
\thicklines
{\color[rgb]{0,0,0}\put(5401,-4201){\line( 1, 0){900}}
\put(6301,-4201){\line( 0,-1){180}}
\put(6301,-4381){\line( 1, 0){630}}
\put(6931,-4381){\line( 0,-1){540}}
\put(6931,-4921){\line(-1, 0){900}}
\put(6031,-4921){\line( 0, 1){180}}
\put(6031,-4741){\line(-1, 0){630}}
\put(5401,-4741){\line( 0, 1){540}}
}%
\thinlines
{\color[rgb]{0,0,0}\put(5041,-5011){\framebox(1350,2160){}}
}%
{\color[rgb]{0,0,0}\put(4681,-4561){\framebox(900,540){}}
}%
{\color[rgb]{0,0,0}\put(4951,-3661){\line( 0,-1){540}}
\put(4951,-4201){\line(-1, 0){180}}
\put(4771,-4201){\line( 0, 1){1080}}
\put(4771,-3121){\line( 1, 0){1350}}
\put(6121,-3121){\line( 0,-1){540}}
\put(6121,-3661){\line(-1, 0){1170}}
}%
\thicklines
{\color[rgb]{0,0,0}\put(4951,-3661){\line( 1, 0){1170}}
\put(6121,-3661){\line( 0, 1){540}}
\put(6121,-3121){\line(-1, 0){1350}}
\put(4771,-3121){\line( 0,-1){900}}
\put(4771,-4021){\line(-1, 0){ 90}}
\put(4681,-4021){\line( 0,-1){540}}
\put(4681,-4561){\line( 1, 0){900}}
\put(5581,-4561){\line( 0, 1){540}}
\put(5581,-4021){\line(-1, 0){630}}
\put(4951,-4021){\line( 0, 1){360}}
}%
{\color[rgb]{0,0,0}\put(6751,-3121){\line( 1, 0){720}}
\put(7471,-3121){\line( 0,-1){540}}
\put(7471,-3661){\line(-1, 0){900}}
\put(6571,-3661){\line( 0, 1){180}}
\put(6571,-3481){\line(-1, 0){720}}
\put(5851,-3481){\line( 0, 1){540}}
\put(5851,-2941){\line( 1, 0){900}}
\put(6751,-2941){\line( 0,-1){180}}
}%
\put(5221,-4921){\makebox(0,0)[b]{\smash{{\SetFigFont{12}{24.0}{\rmdefault}{\mddefault}{\itdefault}{\color[rgb]{0,0,0}$A$}%
}}}}
\put(6616,-4696){\makebox(0,0)[b]{\smash{{\SetFigFont{12}{24.0}{\rmdefault}{\mddefault}{\itdefault}{\color[rgb]{0,0,0}$G$}%
}}}}
\put(5176,-4381){\makebox(0,0)[b]{\smash{{\SetFigFont{12}{24.0}{\rmdefault}{\mddefault}{\itdefault}{\color[rgb]{0,0,0}$E$}%
}}}}
\put(5806,-4516){\makebox(0,0)[b]{\smash{{\SetFigFont{12}{24.0}{\rmdefault}{\mddefault}{\itdefault}{\color[rgb]{0,0,0}$F$}%
}}}}
\put(5446,-3436){\makebox(0,0)[b]{\smash{{\SetFigFont{12}{24.0}{\rmdefault}{\mddefault}{\itdefault}{\color[rgb]{0,0,0}$D$}%
}}}}
\put(7111,-3436){\makebox(0,0)[b]{\smash{{\SetFigFont{12}{24.0}{\rmdefault}{\mddefault}{\itdefault}{\color[rgb]{0,0,0}$B$}%
}}}}
\put(6256,-3301){\makebox(0,0)[b]{\smash{{\SetFigFont{12}{24.0}{\rmdefault}{\mddefault}{\itdefault}{\color[rgb]{0,0,0}$C$}%
}}}}
\end{picture}%
\nop{
 +---D C--B
 |   |||
 |    A
 |  || ||
 +--EF FG
}

\

The sources are $\{A\}$, $\{B,C\}$, $\{D,E\}$ and $\{F,G\}$, the plain maxsets
$\{A,C,D\}$, $\{A,E,F\}$, $\{A,F,G\}$, $\{B,C\}$ and $\{D,E\}$. The following
figure employs arrows to indicate both that a partition has a maxset and that a
maxset induces a partition.

\begin{eqnarray*}
(A|(BC)(DE)|(FG)) & \leftrightarrow & ACD \\
\downarrow && \uparrow \\
AEF & \leftrightarrow & (A|(FG)(DE)|(BC)) \\
&& \downarrow \\
&& AFG \\
&& \updownarrow \\
&& (A(FG)||(DE)(BC)) 
\end{eqnarray*}

Since $A$ is consistent, the tripartitions $(A|(BC)(DE)|(FG))$ and
$(A|(FG)(DE)|(BC))$ have only maxsets containing $A$. In particular, both have
$\{A,C,D\}$ and $\{A,E,F\}$, but only the second has $\{A,F,G\}$. This maxset
induces the partition $(A(FD)||(DE)(BC))$. This partition has $\{A,F,G\}$ as
its only maxset, showing it to be strongly stable.

In terms of the relation,
{} $(A|(FG)(DE)|(BC)) \Rightarrow (A(FD)||(DE)(BC))$
but not the other way around, proving that $\Rightarrow$ is not symmetric even
on weakly stable partitions. Furthermore, both
{} $(A|(BC)(DE)|(FG)) \Rightarrow (A|(FG)(DE)|(BC))$ and 
{} $(A|(FG)(DE)|(BC)) \Rightarrow (A(FD)||(DE)(BC))$ hold, yet
{} $(A|(BC)(DE)|(FG)) \Rightarrow (A(FD)||(DE)(BC))$ does not hold,
proving that the relation is not transitive.~\qed

The definitions of weak and strong stability can be expressed in terms of the
relation $\Rightarrow$ between tripartitions. Indeed, weak stability requires
the partition to have a maxset that induces the partition itself ($X
\Rightarrow X$) while strong stability also requires that no other tripartition
is obtained this way:

\begin{description}

\item[weakly stable:] $X \Rightarrow X$

\item[strongly stable:] $X \Rightarrow Y$ if and only if $X=Y$

\end{description}

Weakly stable tripartitions always exist, but some set of sources do not have
any strongly stable tripartition. This situation can be avoided by relaxing the
second definition. This can be done in at least three different ways, where
$\stackrel * \Rightarrow$ is the transitive closure of $\Rightarrow$,
expressing the existence of a path of tripartitions.

\begin{definition}

A tripartition is {\em mildly stable} if it is weakly stable and one of the
following alternative conditions hold:

\begin{enumerate}

\item if $X \Rightarrow Y$ then $Y \Rightarrow X$

\item if $X \Rightarrow Y$ then $Y \stackrel * \Rightarrow X$

\item if $X \stackrel * \Rightarrow Y$ then $Y \stackrel * \Rightarrow X$

\end{enumerate}

\end{definition}

Rather than insisting on $Y = X$, these conditions allow for $X$ to lead to
some other tripartition $Y$, but only if coming back to $X$ is possible in some
way. In particular, the first requires that a single step suffices, the second
allows for a path, and the third disallows for a tripartition to be reachable
without the original tripartition being reachable from it.

None of the three conditions imply weak stability, since they only constraint
the case where $X \Rightarrow Y$ without saying anything about $X \Rightarrow
X$. This is why the definition of mild stability also requires the tripartition
to be weakly stable.

Conversely, weak stability does not imply any of them: $X \Rightarrow X$ only
implies that a maxset $M$ of $X$ induces $X$ itself, but $X$ may also have
another maxset $M'$ that induces a different partition $Y$. In this case, $X
\Rightarrow Y$ and $X \stackrel * \Rightarrow Y$, which is the situation the
three conditions constraint.

All of them condition the case where $X \Rightarrow Y$; this formula means that
starting from the tripartition $X$ one of the alternative results of merging
(one of the maxsets) leads to the partition $Y$. All three conditions accept
$X$ if $X$ and $Y$ can be somehow be considered ``peer alternatives'': given
that $Y$ is an alternative to $X$, then $X$ is an alternative to $Y$. If the
second does not hold, then $Y$ is an way ``more stable'' than $X$, since the
latter leads to the former but not the other way around.

The second and the third definitions may look the same, but they are not; they
are all different. This will be shown after a general lemma is proved on how to
build counterexamples.

\begin{theorem}

Some sets of sources have no mildly stable partitions according to the first
definition.

\end{theorem}

\proof The counterexample is the following: the sources are
{} $\{A,A',A''\}$,
{} $\{B,B',B''\}$,
{} $\{C,C',C''\}$ and
{} $\{D,D',D''\}$.
For each source, a group of four maxsets contains it all; for the first source,
these are
{} $\{A,A',A'',B,B',C\}$,
{} $\{A,A',A'',C,C',D\}$ and
{} $\{A,A',A'',D,D',B\}$.
Another similar group of four maxsets contains all of $\{B,B',B'\}$, etc. These
maxsets do not contain each other since: a.\ each has a whole source and a part
of the others; and b.\ the ones that contain the same whole source (like two
that contain all of $\{A,A',A''\}$ have the first two formulae of another
source (like $\{B,B'\}$) and one of yet another (like $\{C\}$). Since these
sets of letters do not contain each other, they can be realized by maxsets
thanks to Lemma~\ref{synthesis}.

Since mildly stable tripartitions are also weakly stable by definition, only
tripartitions induced by one of these maxsets can be mildly stable. The first
maxset $\{A,A',A'',B,B',C\}$ induces the following tripartition:

\[
((AA'A'')|(BB'B'')(CC'C'')|(DD'D''))
\]

This tripartition has two maxsets: $\{A,A',A'',B,B',C\}$ and
$\{A,A',A'',C,C',D\}$. The first is a maxset because it induces the partition
itself; the second is an alternative because it contains $C'$ from the second
class, which the first does not. This maxset induces a different partition:

\[
((AA'A'')|(CC'C'')(DD'D'')|(BB'B''))
\]

Again, this tripartition has the maxset $\{A,A',A'',C,C',D\}$ because it was
induced by it, but also $\{A,A',A'',D,D',B\}$ because this maxset contains $D'$
from the second class, which is not in the previous maxset. This maxset induces
the partition:

\[
((AA'A'')|(DD'D'')(BB'B'')|(CC'C''))
\]

This tripartition has the maxset $\{A,A',A'',D,D',B\}$ but also
$\{A,A',A'',B,B',C\}$ because the latter has $B'$ while the first does not.
This was the maxset the proof started from. Therefore, a cycle of three
tripartition has been shown:

\[
\begin{array}{cccc}
&				&((AA'A'')|(BB'B'')(CC'C'')|(DD'D''))	\\
&\hfill\swarrow								\\
((AA'A'')|(CC'C'')(DD'D'')|(BB'B''))&	& \uparrow			\\
&\hfill\searrow								\\
&				&((AA'A'')|(DD'D'')(BB'B'')|(CC'C''))
\end{array}
\]

The point of the proof is that, apart from $\{A,A',A''\}$, each maxset has two
formulae in a source $S$ and one in another $S'$; its induced tripartition has
$S$ and $S'$ in the second class. However, another maxset has two formulae in
$S'$, and is therefore an alternative maxset of the same tripartition. Its
induces tripartition has $S$ in the third class and another source in the
second. A mechanism like this allows building cycles of tripartitions and
maxsets of arbitrary length.

What holds for the maxsets containing $\{A,A',A''\}$ holds for the maxsets
containing $\{B,B',B''\}$, for the maxsets containing $\{C,C',C''\}$ and for
the maxsets containing $\{D,D',D''\}$ because these are the same apart from a
renaming of the formulae. Therefore, this set of sources has no mildly stable
partition according to the first definition.~\qed

The mechanism used in the proof can be generalized so that counterexamples and
specific scenarios can be easily created. This is in the same line of
Lemma~\ref{synthesis}, which shows that sets of letters always express maxsets
given that no set contain another.

For tripartitions, the construction is not that clean. Ideally, it should be
possible to give an arbitrary graph and obtain a set of sources so that their
tripartitions are related as in the graph. Instead, the following lemma shows
how to obtain a graph only for the weakly stable tripartitions, and only if the
graph has a specific form.

\begin{lemma}
\label{realize}

Given a directed graph of $n$ disconnected subgraphs of $n$ nodes each, a set
of sources exists such that the graph expresses its
tripartition--maxset--tripartition relation.

\end{lemma}

\proof Let $X^i_j$ be the $j$-th node of the $i$-th subgraph. The existence of
an edge from $X^i_j$ to $X^i_z$ is denoted by
{} $X^i_j \rightarrow X^i_z$.

The sources are $S_0,S_1,\ldots,S_n$ where $S_i=\{F_i,F_i',F_i''\}$. The first
subgraph is obtained by the following maxsets:

\[
M^0_j = \{F_0,F_0',F_0'',F_j,F_j'\} \cup \{F_z \mid X^i_j \rightarrow X^i_z\}
\mbox{ for every }
j \in [1,n]
\]

The maxsets for the other subgraphs are described below, for the moment the
only important part of the construction is that none of them contain all three
formulae $F_0$, $F_0'$ and $F_0''$.

The weakly stable tripartitions are those induced by at least a maxset. The
tripartition $P_j$ induced by $M^0_j$ has $S_0=\{F_0,F_0',F_0''\}$ in the first
class, $S_j=\{F_j,F_j',F_j''\}$ in the second together with all
$S_z=\{F_z,F_z',F_z''\}$ such that $X^0_j \rightarrow X^0_z$. The remaining
sources are in its third class.

\[
P^0_j = (S_0|S_jS_z\ldots|\ldots)
\]

Since $S_0$ is consistent, only plain maxsets containing all of $S_0$ can be
maxsets of this tripartition $P^0_j$. These are: $M^0_j$, the maxsets $M^0_z$
such that $X^0_j \rightarrow X^0_z$ contain all formulae in the first class,
and the other maxsets $M^0_w$. These three kinds of maxsets are considered in
turn:

\begin{itemize}

\item $M^0_j$ is a plain maxset and it contains all formulae in the first class
and none in the third; by Lemma~\ref{all-none}, it is a maxset;

\item let $z$ be such that $X^0_j \rightarrow X^0_z$; by construction, $M^0_z$
contains all formulae in the first class and $F_z'$; no other maxset contains
all these four formulae; therefore, this one is maximal both in the first class
and in the union of the first and the second;

\item the other plain maxsets $M^0_w$ are the only ones containing all formulae
in the first class of the tripartition; of the second class they may contain
$F_j$, but neither $F_j'$ (because only $M_j$ does) nor $F_z'$ (only $M^0_z$
does); the intersection of this maxset with the first and the second class of
the partition is therefore strictly contained in $M^0_j$.

\end{itemize}

These three points prove that the maxsets of the tripartition $P^0_j$ are
$M^0_j$ and all $M^0_z$ with
{} $X^i_j \rightarrow X^i_z$.
By construction, $M^0_j$ induces $P^0_j$ while $M^0_z$ induces $P^0_z$.
Therefore, every $P^0_j$ is a weakly stable partition, and $P^0_j \Rightarrow
P^0_z$ if and only if $X^0_j \rightarrow X^0_z$.

By symmetry, what holds for index $i=0$ holds for the others. For example, the
maxset $M^1_j$ induces $P^1_j$ for every $j \not= i$. What makes all these
cases disconnected is that $M^i_j$ contains all three formulae
$F_i,F_i',F_i''$, while no other $M^w_j$ contains $F_i''$.

The only caveat to this construction is that for $i=0$ the three formulae
$F_0,F_0',F_0''$ are in all maxsets and are the first class of every partition.
For $i=1$, the same role is taken by $F_1,F_1',F_1''$. Therefore, the node
$X^1_1$ cannot be represented by $P^1_1$ and $M^1_1$, but by $P^1_0$ and
$M^1_0$. In general, for $j \leq i$ the node $X^i_j$ is represented by
$M^i_{j-1}$ and $P^i_{j-1}$. Apart from this change, every node in the graph
has a corresponding tripartition, and the edges correspond to the relation
among tripartitions.~\qed

While the graphs that can be realized by using this lemma have the quite
specific form of $n$ disconnected subgraphs of $n$ nodes each, they are
sufficient for showing some interesting properties. For example, that
$\Rightarrow$ is neither symmetric nor transitive even when restricting to
weakly stable partitions only is proved by a simple graph like $X \Rightarrow
Y$ and $Y \Rightarrow Z$; since it is composed of three nodes, it has to be
replicated three times.

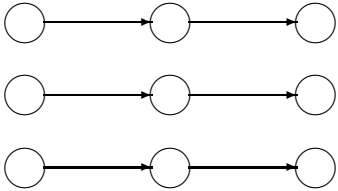
\setlength{\unitlength}{5000sp}%
\begingroup\makeatletter\ifx\SetFigFont\undefined%
\gdef\SetFigFont#1#2#3#4#5{%
  \reset@font\fontsize{#1}{#2pt}%
  \fontfamily{#3}\fontseries{#4}\fontshape{#5}%
  \selectfont}%
\fi\endgroup%
\begin{picture}(1658,938)(4392,-3950)
{\color[rgb]{0,0,0}\thinlines
\put(4501,-3121){\circle{202}}
}%
{\color[rgb]{0,0,0}\put(5221,-3121){\circle{202}}
}%
{\color[rgb]{0,0,0}\put(5941,-3121){\circle{202}}
}%
{\color[rgb]{0,0,0}\put(4591,-3121){\vector( 1, 0){540}}
}%
{\color[rgb]{0,0,0}\put(5311,-3121){\vector( 1, 0){540}}
}%
{\color[rgb]{0,0,0}\put(4501,-3481){\circle{202}}
}%
{\color[rgb]{0,0,0}\put(5221,-3481){\circle{202}}
}%
{\color[rgb]{0,0,0}\put(5941,-3481){\circle{202}}
}%
{\color[rgb]{0,0,0}\put(4591,-3481){\vector( 1, 0){540}}
}%
{\color[rgb]{0,0,0}\put(5311,-3481){\vector( 1, 0){540}}
}%
{\color[rgb]{0,0,0}\put(4501,-3841){\circle{202}}
}%
{\color[rgb]{0,0,0}\put(5221,-3841){\circle{202}}
}%
{\color[rgb]{0,0,0}\put(5941,-3841){\circle{202}}
}%
{\color[rgb]{0,0,0}\put(4591,-3841){\vector( 1, 0){540}}
}%
{\color[rgb]{0,0,0}\put(5311,-3841){\vector( 1, 0){540}}
}%
\end{picture}%
\nop{
O => O => O
O => O => O
O => O => O
}

Three subgraphs of a simple cycle each are the proof that some sources have no
mildly stable partition according to the first definition. Indeed, a set of
sources can be realized so that it has exactly these weakly stable partitions;
for none it holds that $X \Rightarrow Y$ implies $Y \Rightarrow X$; therefore,
these sources have no mildly stable partition according to the first
definition.

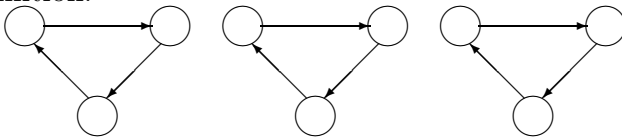
\setlength{\unitlength}{5000sp}%
\begingroup\makeatletter\ifx\SetFigFont\undefined%
\gdef\SetFigFont#1#2#3#4#5{%
  \reset@font\fontsize{#1}{#2pt}%
  \fontfamily{#3}\fontseries{#4}\fontshape{#5}%
  \selectfont}%
\fi\endgroup%
\begin{picture}(3098,668)(4662,-3770)
{\color[rgb]{0,0,0}\thinlines
\put(4771,-3211){\circle{202}}
}%
{\color[rgb]{0,0,0}\put(5131,-3661){\circle{202}}
}%
{\color[rgb]{0,0,0}\put(5491,-3211){\circle{202}}
}%
{\color[rgb]{0,0,0}\put(5851,-3211){\circle{202}}
}%
{\color[rgb]{0,0,0}\put(6211,-3661){\circle{202}}
}%
{\color[rgb]{0,0,0}\put(6571,-3211){\circle{202}}
}%
{\color[rgb]{0,0,0}\put(6931,-3211){\circle{202}}
}%
{\color[rgb]{0,0,0}\put(7291,-3661){\circle{202}}
}%
{\color[rgb]{0,0,0}\put(7651,-3211){\circle{202}}
}%
{\color[rgb]{0,0,0}\put(4861,-3211){\vector( 1, 0){540}}
}%
{\color[rgb]{0,0,0}\put(5446,-3301){\vector(-1,-1){270}}
}%
{\color[rgb]{0,0,0}\put(5086,-3571){\vector(-1, 1){270}}
}%
{\color[rgb]{0,0,0}\put(5941,-3211){\vector( 1, 0){540}}
}%
{\color[rgb]{0,0,0}\put(6526,-3301){\vector(-1,-1){270}}
}%
{\color[rgb]{0,0,0}\put(6166,-3571){\vector(-1, 1){270}}
}%
{\color[rgb]{0,0,0}\put(7021,-3211){\vector( 1, 0){540}}
}%
{\color[rgb]{0,0,0}\put(7606,-3301){\vector(-1,-1){270}}
}%
{\color[rgb]{0,0,0}\put(7246,-3571){\vector(-1, 1){270}}
}%
\end{picture}%
\nop{
O -> O    O -> O    O -> O    
^    |    ^    |    ^    |    
 \   |     \   |     \   |    
   \ V       \ V       \ V    
     O         O         O    
}

This lemma allows for proving that the three definitions of mild stability are
different from each other. A graph suffices, since it can be implemented by
replicating it for a number of times. Since the lemma only concerns weakly
stable partitions, the loops are omitted.

\[
\begin{array}{ccccc}
X & \Leftrightarrow & Y          & \Rightarrow Z & \Rightarrow W \\
  &                 & \Leftarrow
\end{array}
\]

The relation is defined by:
{} $X \Rightarrow Y$,
{} $Y \Rightarrow X$,
{} $Y \Rightarrow Z$,
{} $Z \Rightarrow X$ and
{} $Z \Rightarrow W$.
Since it has four nodes, it has to be replicated in four copies to allow for
the application of the lemma. Only the first copy is considered, since the
others are identical.

The tripartition $X$ is mildly stable according to the first definition, since
$X \Rightarrow Y$ and $Y \Rightarrow X$. The tripartition $Y$ is not, since $Y
\Rightarrow Z$ but not the converse. However, since $Z \Rightarrow X$ and $X
\Rightarrow Y$, it follows $Z \stackrel * \Rightarrow Y$. Therefore $Y$ is
mildly stable according to the second definition but not to the first. Apart
from $W$, none of the tripartitions is mildly stable according to the third
definition since, for example, $Y \stackrel * \Rightarrow W$ but not the other
way around.

This counterexample shows how mildly stable partitions can be visualized: the
first definition requires all outgoing edges to form cycles of length at most
two; the second is the same without constraints on the length; the third can be
recast in terms of maximality.

\begin{theorem}

The mildly stable tripartitions of a set of sources according to the third
definition are the maximal tripartitions with respect to the relation
$\stackrel * \Rightarrow$.

\end{theorem}

\proof By definition, a tripartition $X$ is not maximal if $X \stackrel *
\Rightarrow Y$ for some $Y$, but not the converse. This is exactly the
definition of $X$ being not maximal with respect to $X \stackrel * \Rightarrow
Y$.~\qed

This theorem implies that every set of sources has some mildly stable partition
according to the second definition, since $\stackrel * \Rightarrow$ is by
construction reflexive and transitive, and the number of tripartitions is
finite. The same can be proved for the second definition.

\begin{theorem}

Every set of sources has mildly stable partitions according to the second
definition.

\end{theorem}

\proof The relation $\Rightarrow$ is serial: every tripartition $X$ has at
least a maxset, and this maxset induces a tripartition $Y$; this is the
definition of a serial relation: for every $X$ there exists an $Y$ such that $X
\Rightarrow Y$. Visualizing $\Rightarrow$ as a graph, every node has at least
an outgoing edge.

Some tripartitions may not be mildly stable. This is the case if $X \Rightarrow
Y$ holds but $Y \stackrel * \Rightarrow X$ does not. In both cases, $X$ and all
other tripartitions $Z$ such that $Z \stackrel * \Rightarrow X$ can be removed
from consideration. By assumption, $Y \stackrel * \Rightarrow Z$ does not hold;
such a removal does not change the tripartitions that are reachable from $Y$.

The same argument can be repeated for $Y$: either $Y$ is mildly stable or some
tripartitions can be removed from consideration. Since the number of
tripartitions is finite, at some point either a mildly stable partition is
found or only a single tripartition $W$ is left. In such a case, $W \Rightarrow
W$ and the partition is strongly, and therefore mildly, stable.~\qed

\section{Multiple classes}
\label{general}

Dividing the sources in two classes is based on a very clear-cut distinction: a
source is either reliable (it always provide right information) and unreliable
(may provide wrong information). Also the tripartitions are sharply defined: a
source always, sometimes or never provide right information. Partitions in four
or more classes cannot be based on a similarly uncontroversial division. Yet, a
source providing three correct formulae out of four is more trustable than a
source providing one out of five. The problem is where to place the lines.

Even more general mechanisms can be employed. An high-level summary of what is
done in the previous sections gives at least two possible directions of
generalization. For both bipartitions and tripartitions, the principle is:%
{} a.\ every partition has a set of maxsets; and
{} b.\ every maxset induces a partition.
Abstracting from the specific realization, a partition is a way for
representing the relative reliability of the sources, and a maxset is a way of
merging formulae given the reliability of their sources. Therefore, the
principle is:

\begin{enumerate}

\item from a reliability assessment of the sources, define how they can be
merged;

\item from a possible result of merging, assess the reliability of the sources.

\end{enumerate}

The reliability of the sources is not necessarily a partition; it can for
example be an ordering of the sources, or a numerical evaluation over them. A
possible result of merging could be a maxset, but some integration methods
output an ordering over the interpretations. Various possibilities are explored
in this section.

\subsection{Percentage bounds}

The most obvious extension to the method is to divide the sources on the
percentage of right formulae they provide. This allows for as many classes as
wanted. For example, the first class contains the sources that provided only
correct formulae, the second those providing between 90\% and 100\% of correct
formulae, etc.

A problem with {\em every} method of this kind is that the bounds are
arbitrary: why not setting the division at 89\% instead of 90\%? Why should a
source providing 89\% of correct formulae be in the third class while another
just a little better at 90\% is in the second? This is considered in a further
generalization, for the moment the reader is asked to accept a division based
on some fixed bounds with some intuitive appeal: the reliability bound is fixed
at 90\%, unreliability at 50\%. These numbers consider a source reliable if it
provides 90\% or more of correct formulae, unreliable if it provides less than
50\% of them. The technical results in this section hold for every other fixed
percentages.

Let $C(S_i,M)$ be the fraction of formulae in the set $S_i$ that are consistent
with the set $M$. This function is used with a source $S_i$ and a maxset $M$.
Therefore, $C(S_i,M)$ indicates how reliable the source $S_i$ is assuming that
the state of the word is described by the maxset $M$.

\ttytex{
\[
C(S_i,M)=\frac{|\{A \in S_i \mid M \not\models \neg A\}|}{|S_i|}
\]
}{
           |{A in S_i | M |/= -A}|
C(S_i,M) = -----------------------
                   |S_i|
}

This function allows for evaluating a source given a maxset $M$. In particular,
for the two bounds 90\% and 50\%, it produces the partition of the sources
$(R|P|U)$:

\begin{eqnarray*}
R &=& \{ S_i \mid 0.9 \leq C(S_i,M)		\} \\
P &=& \{ S_i \mid 0.5 \leq C(S_i,M) < 0.9	\} \\
U &=& \{ S_i \mid          C(S_i,M) < 0.5	\}
\end{eqnarray*}

Some limit cases are of interest. Setting both boundaries at 0\% makes $P$ and
$U$ empty and $R$ contain all sources. The only induced partition is that
containing all sources in its first class; this partition is strongly stable;
all plain maxsets are therefore maxsets of a strongly stable partition.

Fixing the boundaries at 100\% and 0\% makes $U$ empty; $R$ and $P$
respectively contain the reliable and unreliable sources as in a bipartitions.
The same holds if both boundaries are 100\%, but this time $P$ is empty and $U$
contains the unreliable sources. As in bipartitions, these are the sources that
provide from zero to all but one correct formulae. The examples on bipartitions
carry to this case. One such examples proves the existence of weakly stable
partitions:

\begin{quote}

for some bounds, strongly stable partitions exists for all set of sources.

\end{quote}

The question is whether strongly stable partitions always exist for an
arbitrary pair of percentages. Actually, even the existence of weakly stable
partition is no longer obvious. Indeed, even if a maxset induces a partition,
it may not be a maxset of it. This was the case for the definitions used in the
previous sections, and many proofs rely on this fact.

Given that reliability does not depend only on the presence or absence of
formulae consistent with a maxset but also on how many these are, the number of
times the same formula is provided by a source matter. To simplify the
notation, a formula $A$ provided nine times by a source is written $A^9$. The
implicit assumption that sources are multisets instead of sets is not really
necessary, as $A^9$ is essentially equivalent to $A \vee T_1$, \ldots, $A \vee
T_9$, where each $T_i$ is a formula with a single model that does not satisfy
any other formula; this may require new variables, forgetting
which~\cite{lang-etal-03} results in nine copies of $A$.

\begin{theorem}

For some percentage bounds, the partition induced by a maxset does not have
that maxset.

\end{theorem}

\proof The claim is proved by a counterexample: the bounds are $90\%$ and
$50\%$, the sources are $\{A^9,C\}$ and $\{C^8,D^2\}$, the plain maxsets are
$\{A,B\}$, $\{A,C\}$, and $\{C,D\}$.

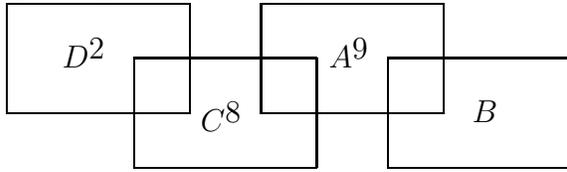
\setlength{\unitlength}{5000sp}%
\begingroup\makeatletter\ifx\SetFigFont\undefined%
\gdef\SetFigFont#1#2#3#4#5{%
  \reset@font\fontsize{#1}{#2pt}%
  \fontfamily{#3}\fontseries{#4}\fontshape{#5}%
  \selectfont}%
\fi\endgroup%
\begin{picture}(2814,834)(3949,-3493)
\thinlines
{\color[rgb]{0,0,0}\put(4591,-3481){\framebox(900,540){}}
}%
{\color[rgb]{0,0,0}\put(5221,-3211){\framebox(900,540){}}
}%
{\color[rgb]{0,0,0}\put(5851,-3481){\framebox(900,540){}}
}%
{\color[rgb]{0,0,0}\put(3961,-3211){\framebox(900,540){}}
}%
\put(5041,-3301){\makebox(0,0)[rb]{\smash{{\SetFigFont{12}{24.0}{\rmdefault}{\mddefault}{\itdefault}{\color[rgb]{0,0,0}$C$}%
}}}}
\put(5671,-2986){\makebox(0,0)[rb]{\smash{{\SetFigFont{12}{24.0}{\rmdefault}{\mddefault}{\itdefault}{\color[rgb]{0,0,0}$A$}%
}}}}
\put(6391,-3256){\makebox(0,0)[rb]{\smash{{\SetFigFont{12}{24.0}{\rmdefault}{\mddefault}{\itdefault}{\color[rgb]{0,0,0}$B$}%
}}}}
\put(5041,-3256){\makebox(0,0)[lb]{\smash{{\SetFigFont{12}{16.8}{\rmdefault}{\mddefault}{\itdefault}{\color[rgb]{0,0,0}$8$}%
}}}}
\put(5671,-2941){\makebox(0,0)[lb]{\smash{{\SetFigFont{12}{16.8}{\rmdefault}{\mddefault}{\itdefault}{\color[rgb]{0,0,0}$9$}%
}}}}
\put(4366,-2986){\makebox(0,0)[rb]{\smash{{\SetFigFont{12}{24.0}{\rmdefault}{\mddefault}{\itdefault}{\color[rgb]{0,0,0}$D$}%
}}}}
\put(4366,-2941){\makebox(0,0)[lb]{\smash{{\SetFigFont{12}{16.8}{\rmdefault}{\mddefault}{\itdefault}{\color[rgb]{0,0,0}$2$}%
}}}}
\end{picture}%
\nop{
+---------+      +----------+ 
| D2   +--+------+--+ A9 +--+-------+
|      |  | C8   |  |    |  |    B  |
+------+--+      +--+----+--+       |
       +------------+    +----------+
} 

The sources have the following percentage of formulae consistent with the
plain maxset $\{A,C\}$:

\begin{itemize}

\item the maxset is consistent with $A$ but not with $B$; the percentage of
consistent formulae of $\{A,B\}$ is therefore 90\%;

\item the maxset is consistent with $C$ but not with $D$, 80\% of $\{C,D\}$.

\end{itemize}

The induced partition is therefore $((A^9B)|(C^8D^2))$. Since $\{A,B\}$ is the
only plain maxset that is consistent with all formulae in the first class of
the partition, it is the only maxset of the partition. Therefore, $\{A,C\}$
induces this partition but is not a maxset of it.~\qed

The proof that every set of sources has a weakly stable partition was based on
the property that the partition induced by a maxset has that maxset. It
therefore no longer works for percentage bounds. The claim still holds, but for
a different reason, suggested by the above proof: a plain maxset is not a
maxset of the induced partition because of another maxset that is ``more
consistent'' with the formulae in the first class of the partition. In turn,
this maxset may not induce a weakly stable partition for the same reason, but
at some point a maximum is reached.

\begin{lemma}

For every pair of bounds there exists an irreflexive, antisymmetric and
transitive ordering $\prec$ among plain maxsets such that, if $M$ is not a
maxset of its induced partition then $M \prec M'$ for some maxset $M$' of this
partition.

\end{lemma}

\proof The bounds define how a maxset $M$ induces a partition $(R|P|U)$. In the
following proof the percentage bounds are set at 90\% and 50\% for the sake of
explanation, but every other pair would work. The ordering is based on the
following function.

\[
e(M) = (M \cap R, M \cap P)
\mbox{ where }
(R|P|U) \mbox{ is the partition induced by } M
\]

The function $e(M)$ does not have the partition as an argument because the
partition is the one induced by $M$. For example, $e(M')$ is defined in terms
of the partition $(R'|P'|U')$ induced by $M'$.

Let $(X,Y) \prec (Z,W)$ if either $X \subset Z$ or $X = Z$ and $Y \subset W$.
The ordering on maxsets is defined by $e(M) \prec e(M')$. This ordering is by
construction irreflexive, symmetric and transitive. Remains to prove that if
$M$ is not a maxset of its induced partition then $e(M) \prec e(M')$ for some
maxset $M'$.

\

Let $M$ be a plain maxset that is not a maxset of its induced partition
$(R|P|U)$. Since $M$ is consistent but not a maxset of $(R|P|U)$, another
consistent set of formulae is strictly larger than $M$ within $R$ or within $R
\cup P$. Such a set can be added some formulae to obtain a maxset $M'$ with the
same property: either
{} $M \cap R \subset M' \cap R$
or
{} $M \cap (R \cup P) \subset M' \cap (R \cup P)$.
These conditions can be rewritten as:

\begin{enumerate}

\item $M \cap R \subset M' \cap R$; or
\label{contain-r}

\item $M \cap R = M' \cap R$ and $M \cap P \subset M' \cap P$.
\label{contain-p}

\end{enumerate}

Let $(R'|P'|U')$ be the partition induced by $M'$. In both Case~\ref{contain-r}
and Case~\ref{contain-p}, $M \cap R \subseteq M' \cap R$. By definition of
induced partition, if $S_i \in R$ the maxset $M$ contains at least 90\% of the
formulae in $S_i$. Since $M \cap R \subseteq M' \cap R$, also $M'$ contains at
least 90\% of $S_i$, implying that $S_i \in R'$. This holds for every source in
$R$; therefore, $R \subseteq R'$.

\


In the first of the two cases above, $M \cap R \subset M' \cap R$. Since $R
\subseteq R'$ it follows that $M \cap R \subset M' \cap R'$. Therefore, $e(M)
\prec e(M')$.

\


In the second case, $M \cap R = M' \cap R$ and $M \cap P \subset M' \cap P$.
Since $R \subseteq R'$, the equality implies $M \cap R \subseteq M' \cap R'$,
which is the first condition for $e(M) \prec e(M')$ to hold.

If a source is in $P$, then $M$ contains between 50\% and 90\% of its formulae.
Since $M \cap P \subset M' \cap P$, then all these formulae are also in $M'$.
But $M'$ may also contain other formulae of this source; therefore, this source
can be in $P'$ but also in $R'$. Formally, $P \subseteq P' \cup R'$.

Let $F$ be a formula in $M' \cap P$ that is not in $M \cap P$, and $S_i$ the
source containing it. Since $F$ is in $P$ but not in $M \cap P$, it is not in
$M$. From $S_i \in P$, it follows $S_i \in R' \cup P'$. If $S_i \in R'$ then
$M' \cap R'$ contains $F$ while $M \cap R$ does not (because $F \not\in M$).
This alone implies $e(M) \prec e(M')$. Otherwise, $F$ is in $P'$. Therefore,
$M' \cap P'$ contains $F$ while $M \cap P$ does not, which is the second
condition for $e(M) \prec e(M')$ to hold.~\qed

This lemma is in a way similar to Lemma~\ref{allp} for bipartitions, in that it
relates a maxset with a maxset of its induced partition. Differently than that
it establishes such a relation only if the first maxset is not a maxset of its
induced partition, and only with a single maxset of the induced partition. It
is however sufficient to prove the existence of weakly stable partitions.

\begin{theorem}

For every pair of bounds, weakly stable partitions exist for every set of
sources.

\end{theorem}

\proof Let $M$ be a plain maxset and $(R|P|U)$ its induced partition. If this
partition has $M$ as a maxset, it is weakly stable and the claim is proved.
Otherwise, by the previous lemma there exists $M'$ such that $e(M) \prec
e(M')$. The same argument can be repeated for $M'$: either its induced
partition has $M'$ as a maxset, or another maxset $M''$ exists such that $e(M')
\prec e(M'')$. Since this relation is irreflexive, antisymmetric and transitive
and the set of maxsets is finite, at some point a maxset $M^*$ is reached such
that $e(M^*) \prec e(M''')$ does not hold for any other maxset $M'''$, proving
that $M^*$ is a maxset of its induced partition.~\qed

\subsection{Ordering}

\def\priority{\mbox{priority}}
\def\reliability{\mbox{reliability}}

A different way to implement the principle of evaluating the reliability of the
sources from a possible result of merging is to use orderings for both. In
particular, the relative reliability of sources can be encoded as an ordering
over sources, and a possible result of merging as a priority order over the
propositional interpretations. Such representation requires a priority order
over interpretations to allow evaluating the reliability of the sources, and an
evaluation of the reliability of the sources to induce one or more priority
orderings over the interpretations.

The reliability of sources can be expressed by a relation $\leq$ and the
priority over interpretations by a relation $\preceq$. When restricted to total
orders, these can be recast in terms of functions from sources and
interpretations to integers. The reliability of a source $S_i$ is denoted
$\reliability(S_i)$, the priority of an interpretation $\priority(I)$. They are
related by:

\begin{description}

\item[reliability $\Longrightarrow$ priority:]
$\priority(I) =
 \min \{ \reliability(S_i) \mid I \models A \mbox{ and } A \in S_i \}$

\item[priority $\Longrightarrow$ reliability:]
$\reliability(S_i) = \max \{ \priority(A) \mid A \in S_i \}$
where
$\priority(A) = \min \{ \priority(I) \mid I \in A\}$.

\end{description}

If an interpretation $I$ is satisfied by a formula in a source of reliability 1
and one in a source of reliability 2, then $I$ has priority 1. In general, an
interpretation is likely being the actual state of the word if a reliable
formula supports it. This explains the minimization.

On the converse, a source providing a formula that is considered untruthful is
assessed as unreliable. The evaluation of both sources and interpretations is
qualitative: the number of formulae supporting an interpretation does not
matter, as well as how many formulae a source provides.

When comparing two reliability orderings expressed in numeric form,
normalization is necessary. Indeed, the orderings in which $I$ is evaluated $0$
in both while $J$ is respectively evaluated $1$ and $2$ may be the same if no
model is evaluated to $1$ in the second. In other words, removing the ``empty
levels'' is necessary when comparing two orderings.

An example is shown on the following sources:

\begin{eqnarray*}
S_1 &=& \{x,y\}			\\
S_2 &=& \{\neg x \wedge y\}
\end{eqnarray*}

A stable ordering is $\reliability(S_1)=\reliability(S_2)=0$. All models of
either of the three formulae have priority $0$, the other model $\{\neg x, \neg
y\}$ has priority $1$. As a result, all three formulae have reliability $0$,
and therefore both sources have reliability $0$ as well. This is the original
reliability ordering, which is therefore stable.

Not all orderings are stable, as shown by $\reliability(S_1)=0$ and
$\reliability(S_2)=1$. The priority of models is
$\priority(\{x,y\})=\priority(\{x,\neg y\})=\priority(\{\neg x,y\})=0$, which
leads to $\reliability(S_2)=0$.

\subsection{Weighted Merge}

The implicit assumption when using maxsets is that merging consists in
collecting as many formulae as possible while retaining consistency. The result
is a subset of the formulae provided by the sources. Semantically, a model of
the merged formulae can only be one satisfying at least one of the formulae;
typically it satisfies more because of the maximality, but at least one has to
be satisfied.

A different approach~\cite{koni-pere-11,koni-lang-marq-02,koni-lang-marq-04} is
based on distances, so that a model may be considered even if it does not
satisfy any formula, but is sufficiently close to all of them.

A number of different definitions have been provided, but often the distance
$d(I,F)$ of a model $I$ to a formula $F$ is the minimal Hamming distance from
$I$ and a model $J$ of $F$. Such a model $I$ is good if the distance is low,
and bad otherwise. The distance is calculated for any of the given formulae
$F_1,\ldots,F_m$; the badness of $I$ is a combination of them, for example a
weighted sum.

\[
d(I,F_1,\ldots,F_m) = w_1 \times d(I,F_1) + \cdots + w_m \times d(I,F_m)
\]

The models at minimal badness form the result of merging. This minimization can
be applied to all models or only those satisfying some integer constraints. The
assumption is that usually some facts are known for certainty, so that a number
of models can be ruled out simply because they conflict with them.

The role the partitions had in the previous sections is now taken by the
weights. They both formalize the reliability of the sources, and therefore the
measure of how much they have to be taken seriously when merging. In the same
way, the possible ways of merging were the maxsets then and are the models now.
The same mechanism of obtaining a possible result of merging from a reliability
assessment, and then to compute a reliability assessment from it, can therefore
be applied.

Let $S_1,\ldots,S_m$ be a set of sources, each comprising one or more formulae.
Each source $S_i$ has a weight $w_i$. When merging, all formulae in a source
$S_i$ have weight $w_i$. This is the same principle of merging with maxsets:
the reliability of a source is taken as a whole, and then every formula it
provides is assumed to have that reliability.

Given the weights, the models at minimal combined distance are defined. This
set of models is the result of merging. If this result is correct, the distance
between it and the formulae in a source allows evaluating the reliability of
the source. Using a model at time or all of them at once allows for obtaining
the definitions of weak and strong stability. The first is defined as follows.

\begin{itemize}

\item merging sources $S_1,\ldots,S_m$ with weights $w_1,\ldots,w_m$ produces
the models at minimal combined weighted distance from the formulae in the
sources;

\item the reliability of a source $S_i$ from a model $I$ is the maximum
distance between $I$ and a formula in $S_i$.

\end{itemize}

The reason for the asymmetry between the two definitions is that the
reliability of a source is not its most truthful formula but its least: a
source is not trustable if it provides some formulae that are very far from the
truth, even if it also provides some that are close. Other way of combining the
distances can be used, such as the average.

The principle of operation is to start with a set of weights, compute the
possible results of merging, and from these compute the distance of the
sources. These should be the same as the original ones, at least qualitatively:
if a source has a weight greater than another, then its resulting distance
should be lower than that of the other.

The following example shows that some weights may be weakly stable while others
are not. This means that the fixpoint definition allows excluding the second
set, and therefore restricting the possible choices.

\begin{eqnarray*}
S_1 &=&	\{A,B\}								\\
S_2 &=&	\{C\}								\\
A &=& 	\neg x_1 \wedge \neg x_2 \wedge \neg x_3 \wedge
	\neg x_4 \wedge \neg x_5 \wedge \neg x_6 \wedge
	\neg x_7							\\
B &=& 	     x_1 \wedge      x_2 \wedge \neg x_3 \wedge
	\neg x_4 \wedge \neg x_5 \wedge \neg x_6 \wedge
	\neg x_7							\\
C &=& 	     x_1 \wedge      x_2 \wedge      x_3 \wedge
	     x_4 \wedge      x_5 \wedge      x_6 \wedge
	     x_7							\\
K &=& 	     x_1 \wedge      x_2 \wedge \neg x_3 \wedge
	\neg x_4 \wedge	\neg x_5 \wedge
	\neg x_7							\\
I &=&	\{ x_1, x_2, x_3, x_4, x_5, \neg x_6, \neg x_7 \}		\\
J &=&	\{ x_1, x_2, x_3, x_4, x_5,      x_6, \neg x_7 \}		\\
\end{eqnarray*}

The formula $K$ expresses the constraints: only its two models $I$ and $J$ are
considered when computing the result of merging. The distance between these
models and the formulae are:

\begin{eqnarray*}
d(I,A) &=& 4	\\
d(I,B) &=& 2	\\
d(I,C) &=& 2	\\
d(J,A) &=& 5	\\
d(J,B) &=& 3	\\
d(J,C) &=& 1
\end{eqnarray*}

The starting point of the
method is to assume some weights for the sources, for example $w_1=w_2=1$; this
is an obvious choice: in lack of information indicating which source to trust
more, they are trusted the same.

Using these weights, the badness of $I$ is
{} $w_1 \times (d(I,A) + d(I,B)) + w_2 \times d(I,C) =
{} 1 \times (4 + 2) + 1 \times 2 = 8$.
The badness of $J$ is
{} $1 \times (5 + 3) + 1 \times 1 = 9$.
The result of merging is therefore $I$ only. This allows for an easy evaluation
of the sources: $4$ for $S_1$ and $3$ for $S_2$. These reliability measures are
different, contradicting $w_1 = w_2$. The original choice of weights is not
stable.

A different outcome is obtained by $w_1 = 1$ and $w_2 = 2$. The badness of $I$
is
{} $1 \times (4+2) + 2 \times 2 = 10$
and that of $J$ is
{} $1 \times (5+3) + 2 \times 1 = 8 + 2 = 10$.
Both models are in the result of merging, so both have to be considered when
checking stability. The distance between $I$ and $S_1$ is $4$, between $I$ and
$S_2$ is $2$. These compare the opposite of $w_1$ and $w_2$, as required
(large weight means high reliability, which in turns should produce a lower
distance). This proves that the weights are weakly stable. A similar
calculation for $J$ shows that the same happens for it, making the weights
strongly stable.

\section{Directions}
\label{directions}

The principle studied in this article is that some reliability orderings can be
excluded on the ground that using them for revising leads to a different
ordering. The way this principle is implemented is by assuming sources to
generate a single or a set of formulae, that the reliability ordering is
represented by a partition of the sources or by weights, and that merging is
done by maxsets or by a weighted sum.

Apart from how the reliability ordering and the merging process are realized,
the principle itself can be extended in many ways.

\subsection{Sure information}

This is possibly the direction of most interest: some sources of unknown
reliability provide formulae to be merged, but some information is also known
for certain. This is a common scenario in belief merge, encoded by integrity
constraints~\cite{koni-pere-99} as follows: only the models that satisfy them
are considered when selecting the ones that make the result of merging.

The fixpoint definition allows for a smooth encoding of formulae that are
considered certain. They are collected in a single source, and only reliability
orderings having that source is in the first class are considered. Since the
first class of an induced partition contains only formulae that are consistent
with the result of the merge, this makes such result to be consistent with all
these formulae.

Formalizing the certain information as a source has the advantage of an uniform
treatment of it along with the other formulae. There is no prior reason why
another source could not be equally certain; if so, it is in the first class of
the reliability partition, together with the source encoding the sure formulae.
In fact, the integrity constraints come from somewhere, the only difference
with the other formulae being that their sources can never be considered
unreliable, not even partially so.

The reason why this variant may be the case of most interest stems from having
certain information to rule out some reliability orderings. The scheme with no
integrity constraint uses only the regular sources to reckon their reliability
ordering. The integrity constraints ground this evaluation on information that
is known for certain.

\subsection{Partial reliability ordering}

Limiting to reliability orderings with a given source in the first class is
only a particular case of partial information over the reliability ordering. If
$S_0$ is such a source, $S_0 \leq S_i$ holds for every other source $S_i$.

More generally, an arbitrary partial knowledge of the ordering can be assumed.
For example, it may be that only $S_1 < S_2$ and $S_2 \leq S_3$ are known. This
rules out the partition $(S_2|S_1S_3)$ but neither $(S_1|S_2S_3)$ nor
$(S_1|S_2|S_3)$. Since bipartitions and tripartitions set a strong division
between classes (all true, some true, none true), constraints of type $\leq$
are better suited for them, as otherwise an apparently harmless statement like
``$S_1$ is strictly more reliable than $S_2$'' would turn ``$S_1$ is only
partly reliable'' into the strong implication that $S_2$ is totally unreliable.
Such strict constraints like $S_1 < S_2$ are better left to multiple-classes
partitions, where they only state that $S_2$ provides fewer correct formulae
than $S_1$.

\subsection{Untrustable sources}

Tripartitions class sources in three groups: totally reliable, partly reliable
and unreliable. The first two are clearly defined, as they are the sources that
provide only correct formulae and some correct formulae. However, the
meaning of the third class changes depending on the considered direction:

\begin{description}

\item[partition $\rightarrow$ maxset] formulae in the third class are included
if consistent with the formulae of the first two;

\item[maxset $\rightarrow$ partition] a source goes in the third class if all
its formulae are inconsistent with the maxset.

\end{description}

In the first point formulae are included only if possible, in the second they
are not. Only if the partition is stable the two coincide. This highlights the
ambiguity of defining a source ``totally unreliable''. It may be a source that
is less reliable than all others but in principle may provide correct
information, but it may also be a source that always provide wrong information.

The third class can therefore be split in two. A maxset may include formulae in
the first subclass if consistent with the rest of it, but does not include any
of the second, simply because these are known being false. In a way, this
fourth class is specular to the first: all its formulae contradict the truth
rather than being consistent with it. Such a class can formalize scenarios in
which a source intentionally tries to lead the revision process in a certain
direction; this raises the question of detecting strategies such as providing
some right information to escape the classification in the fourth class.

\subsection{Flexible bounds}

Dividing the sources on $100\%$ and $0\%$ of correct formulae, as in
bipartition and tripartitions, appeals to a qualitative notion of reliability
and unreliability, every other bound (like $80\%$ or $50\%$) is difficult to
motivate over others (like $90\%$ and $51\%$).

A possible solution is to assume that a partition like $(S_1|S_2S_3|S_4)$ only
means that there exists a reason to assume that $S_1$ is more reliable than
both $S_2$ and $S_3$, which are both more reliable than $S_4$, without
requiring two fixed borders between them. As an example, this partition is
acceptable if the four sources provide $80\%$, $72\%$, $69\%$ and $30\%$
correct formulae. The bound can be fixed at $75\%$ and $50\%$ to support this
division, but also at $79\%$ and $60\%$, or at $79\%$ and $40\%$.

This mechanism lifts the decision of an exact pair of percentages of correct
formulae that define a source reliable or partly reliable. A partition like
$(S_1|S_2S_3|S_4)$ is then stable if such a pair makes the partition itself
induced from the result of merging.

\subsection{Technical results}

Some other possible expansions of this work regards the study of the existing
definitions, rather than the extensions of them. Lemma~\ref{realize} allows for
the creation of a scenario with given partitions and maxsets, but only if these
satisfy certain conditions. It would be interesting to find out whether these
constraints could be lifted.

The two algorithms presented in this article cover the case of bipartitions
only. Extending them to tripartitions or even to the case of arbitrary
partitions would make them applicable to more general situations. Also, the
complexity of inferring from the maxsets of the stable partitions, or checking
for the existence of strongly stable tripartitions, is also an open problem.

\section{Conclusions}

The principle of excluding some reliability orderings when the merging result
they produce conflicts with them can be implemented in many ways. In the
simplest case, sources are classified as reliable and unreliable. The second
group can be split into the partly reliable and unreliable sources.

The change has a big impact on the technical results: while strongly stable
bipartitions always exist, this is not the case for strongly stable
tripartitions. This motivates the definition of an ordering among weakly stable
partitions in order to obtain a form of stability that is more grounded than
weak stability.

Switching to classes based on percentages makes another important property
fail: the partition induced by a maxset may not have that maxset. The existence
of weakly stable partitions for arbitrary pairs of bounds, which was a simple
consequence of this fact, had to be proved by a different argument.

All of this indicates that the technical consequences are strongly dependent on
the details of the definition employed. While the general principle is always
the same (fix a reliability ordering, merge using it, check if the result
agrees with the reliability ordering), the properties of the framework depend
on how it is implemented.

However, some general facts appear to hold regardless:

\begin{enumerate}

\item weakly stable partitions always exist; this is not obvious from the
definition, rather the opposite: a definition of partitions and merge that does
have this property is unusable since weak stability is the most basic
requirement for merging; if some sources do not have such a partition, they
cannot be merged; it is hard to imagine a scenario where information cannot be
integrated, not even with an void result;

\item since maxsets induce partitions and partitions have maxsets, an ordering
of the partitions (or maxsets, or both) always exists; such an ordering can be
used to define a form of maximality of partitions, or at least to exclude
partitions that are strictly dominated by others; the rationale is that if a
reliability ordering lead to a result of merging that implies a different
reliability ordering, the latter is more grounded than the former;

\item as in the general problem of belief revision, the aim is at the same time
to obtain some merging result but also to restrict the possible alternatives as
much as possible, since this leads to a resulting formula that carries as much
information as possible; as a result, whichever pair of definitions are used
for the result of merging and the induced reliability ordering, some stable
partitions should always exists, but as few as possible; this is why
definitions that lead to cases without weakly stable partitions are ruled out;
at the same time, an inflation of weakly stable partitions (like when the lead
to all plain maxsets) motivates the introduction of strong stability or at
least the restriction to maximal weakly stable partitions.

\end{enumerate}

Few articles in the literature are related to the research presented in this
article. This is because most work in belief revision and merging is about
going from a set of formulae to their aggregation. An handful of articles
consider the reverse problem:
Haret, Mailly and Woltran~\shortcite{hare-mail-wolt-16} divide a formula into
simpler formulae that give the original one when merged;
Booth and Nittka~\shortcite{boot-nitt-08} and Liberatore~\shortcite{libe-15}
use the history of previous revision to obtain reliability information;
Liberatore~\shortcite{libe-16} use merging samples to the same aim.
All this research share the reversal of the traditional role of the sources and
the result of merge, where the first are input and the second is output. Apart
from the first cited article, which aims at representing knowledge in some
simple form, the other do this to obtain reliability information, which can be
then used in the following merging. The present article does the same, but does
not assume any prior knowledge. Rather, the role of history or examples is
taken by the merging result itself, which allows to evaluate the sensibility of
the reliability ordering initially assumed.

\bibliographystyle{plain}

\begin{thebibliography}{10}

\bibitem{adam-etal-08}
L.~Adamic, J.~Zhang, E.~Bakshy, and M.S. Ackerman.
\newblock Knowledge sharing and {Y}ahoo {A}nswers: everyone knows something.
\newblock In {\em Proceedings of the Seventeenth International Conference on
  World Wide Web (WWW'08)}, pages 665--674, 2008.

\bibitem{baru-thom-hass-12}
A.~Barua, S.W. Thomas, and A.E. Hassan.
\newblock What are developers talking about? {A}n analysis of topics and trends
  in {S}tack {O}verflow.
\newblock {\em Empirical Software Engineering}, 19(3):619--654, 2012.

\bibitem{boot-nitt-08}
R.~Booth and A.~Nittka.
\newblock Reconstructing an agent's epistemic state from observations about its
  beliefs and non-beliefs.
\newblock {\em Journal of Logic and Computation}, 18:755--782, 2008.

\bibitem{cevo-14}
G.~Cevolani.
\newblock Truth approximation, belief merging, and peer disagreement.
\newblock {\em Synthese}, 191(11):2383--2401, 2014.

\bibitem{dalf-15}
S.~D'Alfonso.
\newblock Belief merging with the aim of truthlikeness.
\newblock {\em Synthese}, pages 1--22, 2015.

\bibitem{delg-dubo-lang-06}
J.P. Delgrande, D.~Dubois, and J.~Lang.
\newblock Iterated revision as prioritized merging.
\newblock In {\em Proceedings of the Tenth International Conference on
  Principles of Knowledge Representation and Reasoning, (KR-2006)}, pages
  210--220, 2006.

\bibitem{ever-koni-marq-15}
P.~Everaere, S.~Konieczny, and P.~Marquis.
\newblock Belief merging versus judgment aggregation.
\newblock In {\em Proceedings of the Fourteenth International Conference on
  Autonomous Agents and Multiagent Systems (AAMAS'15)}, pages 999--1007, 2015.

\bibitem{hare-mail-wolt-16}
A.~Haret, J.-G. Mailly, and S.~Woltran.
\newblock Distributing knowledge into simple bases.
\newblock Technical Report abs/1603.09511, Computing Research Repository
  (CoRR), 2016.

\bibitem{herz-pozo-schw-14}
A.~Herzig, P.~Pozos-Parra, and F.~Schwarzentruber.
\newblock Belief merging in dynamic logic of propositional assignments.
\newblock In {\em Proceedings of the eighth International Symposium on the
  Foundations of Information and Knowledge Systems (FoIKS 2014)}, pages
  381--398. Springer, 2014.

\bibitem{koni-lang-marq-02}
S.~Konieczny, J.~Lang, and P.~Marquis.
\newblock Distance-based merging: a general framework and some complexity
  results.
\newblock In {\em Proceedings of the Eighth International Conference on
  Principles of Knowledge Representation and Reasoning (KR~2002)}, pages
  97--108, 2002.

\bibitem{koni-lang-marq-04}
S.~Konieczny, J.~Lang, and P.~Marquis.
\newblock {DA}$^2$ merging operators.
\newblock {\em Artificial Intelligence}, 157(1-2):49--79, 2004.

\bibitem{koni-pere-11}
S.~Konieczny and R.P. P\'erez.
\newblock Logic based merging.
\newblock {\em Journal of Philosophical Logic}, 40(2):239--270, 2011.

\bibitem{koni-pere-99}
S.~Konieczny and R.~Pino~P\'erez.
\newblock Merging with integrity constraints.
\newblock In {\em Proceedings of the Fifth European Conference on Symbolic and
  Quantitative Approaches to Reasoning and Uncertainty, (ECSQARU'99)}, pages
  233--244, 1999.

\bibitem{lang-etal-03}
J.~Lang, P.~Liberatore, and P.~Marquis.
\newblock Propositional independence - formula-variable independence and
  forgetting.
\newblock {\em Journal of Artificial Intelligence Research}, 18:391--443, 2003.

\bibitem{libe-15}
P.~Liberatore.
\newblock Revision by history.
\newblock {\em Journal of Artificial Intelligence Research}, 52:287--329, 2015.

\bibitem{libe-16}
P.~Liberatore.
\newblock Belief merging by examples.
\newblock {\em {ACM} Transactions on Computational Logic}, 17(2), 2016.

\bibitem{libe-scha-98-b}
P.~Liberatore and M.~Schaerf.
\newblock Arbitration (or how to merge knowledge bases).
\newblock {\em {IEEE} Transactions on Knowledge and Data Engineering},
  10(1):76--90, 1998.

\bibitem{naum-etal-06}
F.~Naumann, A.~Bilke, J.~Bleiholder, and M.~Weis.
\newblock Data fusion in three steps: Resolving schema, tuple, and value
  inconsistencies.
\newblock {\em {IEEE} Data Eng. Bull.}, 29(2):21--31, 2006.

\bibitem{nebe-92}
B.~Nebel.
\newblock {\em Syntax-Based Approaches to Belief Revision}, pages 52--88.
\newblock Cambridge University Press, 1992.

\bibitem{nebe-98}
B.~Nebel.
\newblock How hard is it to revise a belief base?
\newblock In D.~Dubois and H.~Prade, editors, {\em Belief Change - Handbook of
  Defeasible Reasoning and Uncertainty Management Systems, Vol. 3}. Kluwer
  Academic, 1998.

\bibitem{porc-06}
Porchesia.
\newblock {\tt http://wikimedia.7.x6.nabble.com/Porchesia-td962303.html}, 2006.

\bibitem{posn-etal-12}
D.~Posnett, E.~Warburg, P.T. Devanbu, and V.~Filkov.
\newblock Mining {S}tack {E}xchange: expertise is evident from initial
  contributions.
\newblock In {\em Prooceding of the 2012 International Conference on Social
  Informatics (SocialInformatics)}, pages 199--204, 2012.

\bibitem{rott-93}
H.~Rott.
\newblock Belief contraction in the context for the general theory of rational
  choice.
\newblock {\em Journal of Symbolic Logic}, 58(4):1426--1450, 1993.

\bibitem{vieg-etal-07}
F.B. Vi\'egas, M.~Wattenberg, J.~Kriss, and F.~van Ham.
\newblock Talk before you type: coordination in {W}ikipedia.
\newblock In {\em Proceeding of the Fortieth Hawaii International International
  Conference on Systems Science {(HICSS-40})}, page~78, 2007.

\bibitem{wile-gurr-09}
D.~Wiley and S.~Gurrell.
\newblock A decade of development.
\newblock {\em Open Learning: The Journal of Open, Distance and e-Learning},
  24:11--21, 2009.

\end{thebibliography}

\end{document}